\providecommand\FormatOverride{neurips}           %
\providecommand\LogosOverride{camlsys-cambridge-flower-zib-cnrs-sofair}  %
\providecommand\TitleBgOverride{camlsys}          %
\documentclass{paper}

\input{preamble/packages}   %
\input{preamble/macros}     %
\input{preamble/adapter}    %
\input{preamble/logos}      %

\renewcommand{\paperTitleFont}{\fontsize{17.8}{20.5}\selectfont}
\paperTitle{\texorpdfstring{\method: Scaling Local-Update Training\\with Zero-Shot Hyperparameter Transfer}{DistP: Scaling Local-Update Training with Zero-Shot Hyperparameter Transfer}}
\paperTitleRunning{\method: Scaling Local-Update Training with Zero-Shot Hyperparameter Transfer}
\paperTitleLogo{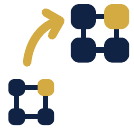}
\paperTitleLogoHeight{2.8em}

\addAffiliation{cam}{University of Cambridge}
\addAffiliation{sofair}{Science of Fundamental AI Research Lab (SOFAIR)}
\addAffiliation{zib}{Zuse Institute Berlin (ZIB)}
\addAffiliation{paris}{CNRS - ENS, PSL Research University, INRIA, Paris, France}
\addAffiliation{flower}{Flower Labs}

\addAuthor{Andrej Jovanović}{equal,cam,sofair}
\addAuthor{Alex Iacob}{equal,cam,sofair}
\addAuthor{Philipp Wiesner}{cam,sofair}
\addAuthor{Christophe Roux}{zib,paris}
\addAuthor{Max Zimmer}{zib}
\addAuthor{Lorenzo Sani}{cam,sofair,flower}
\addAuthor{Sebastian Pokutta}{zib}
\addAuthor{Nicholas D. Lane}{cam,sofair,flower}

\correspondingAuthor{Andrej Jovanović}{aj693@cam.ac.uk}

\begin{document}
\etocdepthtag{main}

\makepapertitle

\begingroup
\setlength{\parskip}{3pt}
\section{Introduction}
\begingroup
\setlength{\parskip}{3pt}
Efficient training of foundation models requires distributing computation across many accelerators~\citep{narayanan2021efficient}. In standard Distributed Data Parallelism (\ddp), gradients are synchronized before every optimization step, creating a bottleneck on low-bandwidth networks. The community has leveraged \emph{local-update training} to reduce this overhead by allowing workers to train model copies independently for several steps before combining their updates~\citep{douillard2023diloco,iacob2025mtdao,jovanovic2026lordo}. At synchronization, a shared model update is obtained by averaging these updates~\citep{mcmahan2017fedavg,stich2019local} or by applying an outer optimizer step~\citep{wang2020slowmo,reddi2021adaptive}.

Although these local-update methods can match the training performance of \ddp
while reducing wall-clock time in low-bandwidth environments~\citep{charles2025dilocoscaling,iacob2025mtdao}, they expand
the hyperparameter search space and increase the cost of tuning.
Beyond the learning rate and batch size, practitioners must choose the worker count, the number of local steps, and the outer optimizer and its hyperparameters.
Moreover, the best settings depend on the training configuration, as
changing the worker count or synchronization interval can shift the
optimal hyperparameters~\citep{charles2025dilocoscaling}. This leads to configuration-specific tuning for these regimes, where the cost
grows with model size~\citep{therien2025muloco}.

A potential solution lies with zero-shot hyperparameter transfer, which enables the reuse of settings tuned on a small, inexpensive reference configuration at larger scales, reducing retuning. Existing work on this topic establishes the transfer rules, albeit for \ddp:
\mup~\citep{yang2022tensorprograms} addresses width, \completep~\citep{dey2025completep} extends transfer to depth, and \completedp~\citep{mlodozeniec2025completedp} incorporates batch size and token budget.
These parameterizations also control how learned features change with scale, supporting better model quality.
However, given that local updates introduce new optimization dynamics, the extension of zero-shot hyperparameter transfer to these regimes remains unclear. As such, we ask:
\begin{list}{}{\setlength{\leftmargin}{3.5em}\setlength{\rightmargin}{3.5em}%
 \setlength{\topsep}{3pt}\setlength{\parsep}{0pt}}
\item[]
\emph{How can zero-shot hyperparameter transfer be extended to local-update training?}
\end{list}
\par\endgroup

We introduce \method, a scaling framework for local-update training. We establish theoretical conditions under which width and depth scaling applied on each worker is preserved under averaging and linear outer optimizers, such as \nesterov momentum, which form weighted sums of current and past updates. For adaptive outer optimizers, such as \adam, which normalize updates using running moment estimates, we derive additional scaling corrections.
We then show why larger local batches and additional workers lead to different learning-rate scaling rules, and when the same local learning rate can be used across synchronization intervals. 

We evaluate \method in language-modeling experiments across model scales, batch sizes, worker counts, local steps, and outer optimizers. Our findings demonstrate that \method improves model quality and reduces retuning in communication-efficient training:
\introtakeawaybox{
\begin{takeawayitems}
\item \textbf{Width and depth scaling transfers across outer optimizers.}
We establish conditions for preserving width and depth scaling~\citep{yang2022tensorprograms,dey2025completep} under linear outer updates and demonstrate transfer empirically.
We derive corrections that make outer \adam competitive with averaging and \nesterov.
\item
\textbf{Batch scaling under averaging has two regimes.}
We show that, under averaging with inner \adam, larger local batches favor square-root learning-rate scaling, whereas additional workers favor the reference rate. We transfer the local rate unchanged across synchronization intervals.
\item \textbf{Outer momentum couples learning rates.} We find that changing batch size or local steps with outer momentum can require joint inner--outer tuning, consistent with~\citet{khaled2025outer}.
\item \textbf{Transfer becomes more beneficial at larger model scales.}
We achieve lower loss at matched compute, with fitted laws suggesting larger gains at higher budgets.
We transfer small-model settings to $823$M parameters, reducing validation loss by \(8.1\%\) and improving mean downstream accuracy by \(10\%\) relative to baselines, without target-scale tuning.
\end{takeawayitems}
}

\section{Preliminaries} \label{sec:distributed_training_preliminaries}
In this section, we provide an overview of local-update training and zero-shot hyperparameter transfer to assist in the exposition of \method. We study distributed training in which workers optimize local model
copies before combining their updates into a shared model.
We call these stages \emph{inner optimization} and
\emph{outer optimization}, respectively.
We consider \(M\) workers jointly minimizing
\(f(\theta)=\frac{1}{M}\sum_{m=1}^{M}f_m(\theta)\), where
\(\theta\) denotes model parameters and \(f_m\) is worker \(m\)'s local loss.
Each worker performs \(K\) local steps of training, indexed by \(k\),
followed by a synchronization step that produces a shared model update.
We write \(\theta_t\) for the shared parameters given to each worker
at round \(t\), and \(\theta^m_{t,k}\) for worker \(m\)'s
local parameters after \(k\) steps of local training.
Following \desloc~\citep{iacob2025desloc}, we use a common framework
to describe parameter and optimizer-state updates.

\paragraph{Local updates.}
Assume that workers start from the shared model checkpoint \(\theta^m_{t,0}=\theta_t\).
At each local step \(k=0,\ldots,K-1\), a worker \(m\) computes a minibatch
gradient \(g^m_{t,k}\) at \(\theta^m_{t,k}\).
Its local (inner) optimizer updates the parameters and optimizer state
\(s^m_{t,k}\) according to its hyperparameters:
\begin{equation}
 (\theta^m_{t,k+1},s^m_{t,k+1})
 =\operatorname{InnerOpt}(\theta^m_{t,k},s^m_{t,k},g^m_{t,k}).
 \label{eq:inner-map}
\end{equation}

\paragraph{Shared model update.}
After \(K\) steps, the average pseudo-gradient is computed:
\begin{equation}
 \Delta_t=\frac{1}{M}\sum_{m=1}^{M}\Delta_t^m,
 \qquad
 \Delta_t^m=\theta_t-\theta^m_{t,K}.
 \label{eq:aggregation}
\end{equation}
Following the \fedopt framework~\citep{reddi2021adaptive}, an outer
optimizer with state \(s_t^{\mathrm{out}}\) computes the shared update:
\begin{equation}
 (\theta_{t+1},s_{t+1}^{\mathrm{out}})
 =\operatorname{OuterOpt}(\theta_t,s_t^{\mathrm{out}},\Delta_t),
 \label{eq:outer-map}
\end{equation}
which determines the new model at each worker \(\theta^m_{t+1,0}=\theta_{t+1}\). Choosing gradient descent with unit learning rate recovers plain
averaging, \(\theta_{t+1}=\theta_t-\Delta_t\)~\citep{mcmahan2017fedavg}.
Other outer optimizers, such as \nesterov and \adam, apply momentum
or adaptive rescaling to the averaged pseudo-gradient \(\Delta_t\).

\paragraph{Local optimizer-state synchronization.}
Optimizer states can synchronize at a different interval from model
parameters~\citep{iacob2025desloc}, or remain
local~\citep{douillard2023diloco,therien2025muloco}.
The convergence guarantees in~\citep{cheng2025convergence,iacob2025desloc}
assume periodic optimizer-state synchronization. We follow \citet{cheng2025convergence} and synchronize parameters and optimizer states every \(K\) local steps, averaging states as:
\begin{equation}
 s^m_{t+1,0}=\frac{1}{M}\sum_{j=1}^{M}s^j_{t,K}.
 \label{eq:broadcast-map}
\end{equation}
The shared parameters and averaged states initialize each worker after synchronization.

\paragraph{Hyperparameter transfer frameworks.}
\label{sec:model-scale-transfer}
Zero-shot hyperparameter transfer aims to reuse settings tuned on a small reference configuration at larger scales without retuning.
In \ddp, transfer rules specify how to adjust weight initialization, rescale layer outputs in the forward pass, and change optimizer hyperparameters to account for differences in the training configuration.
\mup~\citep{yang2022tensorprograms} provides these rules for model width, \completep~\citep{dey2025completep} extends them to depth, and \completedp~\citep{mlodozeniec2025completedp} incorporates batch size and token budget.
The adjustments use scale ratios between the target and reference configurations (denoted by subscript \(0\)):
\(\widthaxis{m_N=N/N_0}\), \(\depthaxis{m_L=L/L_0}\), \(\batchaxis{m_B=B/B_0}\), and \(\dataaxis{m_D=D/D_0}\),
where \(N\) denotes width, \(L\) depth, \(B\) tokens per global batch, and \(D\) the token budget.
For example, with the residual-depth exponent \(\alpha=1\), used in our main experiments,
the inherited rules scale residual branch outputs by
\(\depthaxis{m_L^{-1}}\) and hidden-matrix \adam learning rates by
\(\widthaxis{m_N^{-1}}\sqrt{\batchaxis{m_B}/\dataaxis{m_D}}\).
We refer readers to the cited works for the full prescriptions.
Our central question is how these frameworks extend to local-update
training, which we discuss in \Cref{sec:new-distributed-training-scaling}.

\begin{widefigure}[t]
    \centering
    \resizebox{\linewidth}{!}{%
\begin{tikzpicture}[
  x=1pt,y=1pt,
  font=\fontsize{8}{9}\selectfont\rmfamily,
  overviewbox/.style={draw=ThemeNavy!60,fill=ThemeNavy!3,
    rounded corners=3pt,line width=.55pt,inner sep=2pt},
  workerbox/.style={overviewbox,fill=ThemeNavy!8,minimum width=42pt,
    minimum height=14pt,font=\fontsize{8}{9}\selectfont\bfseries},
  innerbox/.style={overviewbox,draw=ThemeNavy!65,fill=ThemeNavy!5,
    minimum width=59pt,minimum height=14pt},
  stepbox/.style={overviewbox,draw=CalloutGold,fill=CalloutPeach,
    minimum width=33pt,minimum height=14pt},
  flow/.style={-{Latex[length=3pt,width=3pt]},line width=.65pt},
  workerflow/.style={flow,draw=ThemeNavy}
]
  \node[overviewbox,minimum width=84pt,minimum height=82pt]
    (dimensions) at (42,0) {};
  \node[anchor=north west,inner sep=0pt,
    font=\fontsize{8}{9}\selectfont\bfseries]
    at (5,35) {Small reference};
  \node[overviewbox,draw=ScaleWidth,fill=ScaleWidth!6,
    minimum width=70pt,minimum height=14pt,inner sep=1pt] (widthrule)
    at (42,14) {\mup\ \textcolor{ScaleWidth}{\(N\)}};
  \node[overviewbox,draw=ScaleDepth,fill=ScaleDepth!6,
    minimum width=70pt,minimum height=14pt,inner sep=1pt] (depthrule)
    at (42,-8) {\completep};
  \node[overviewbox,draw=ScaleTokens,fill=ScaleTokens!6,
    minimum width=70pt,minimum height=14pt,inner sep=1pt] (trainingrule)
    at (42,-30) {\completedp};
  \draw[flow,draw=ScaleDepth] (widthrule.south) --
    node[right,inner sep=1pt,font=\fontsize{7}{8}\selectfont]
    {\textcolor{ScaleDepth}{\(+L\)}} (depthrule.north);
  \draw[flow] (depthrule.south) --
    node[right,inner sep=1pt,font=\fontsize{7}{8}\selectfont]
    {\(+\textcolor{ScaleBatch}{b},\textcolor{ScaleTokens}{D}\)} (trainingrule.north);

  \node[overviewbox,minimum width=20pt,minimum height=21pt,
    right=9pt of dimensions] (theta) {\(\theta_t\)};
  \node[workerbox] (workerone) at (148,24) {Worker \(1\)};
  \node[workerbox,below=6pt of workerone] (workertwo) {Worker \(2\)};
  \node[workerbox,below=17pt of workertwo] (workerm) {Worker \(M\)};
  \foreach \suffix in {one,two,m} {
    \node[innerbox,right=8pt of worker\suffix] (inner\suffix)
      {\(\operatorname{InnerOpt}(\eta,\epsilon)\)};
    \node[stepbox,right=8pt of inner\suffix] (steps\suffix)
      {\(K\) steps};
    \draw[workerflow] (worker\suffix.east) -- (inner\suffix.west);
    \draw[workerflow] (inner\suffix.east) -- (steps\suffix.west);
  }
  \foreach \column in {worker,inner,steps} {
    \node[inner sep=0pt] at ($(\column two.south)!.5!(\column m.north)$)
      {\(\vdots\)};
  }
  \coordinate (broadcast) at ($(workerone.west)+(-12pt,0)$);
  \draw[line width=.65pt] (theta.east) -- (theta.east -| broadcast);
  \foreach \suffix in {one,two,m} {
    \draw[flow] (theta.east -| broadcast) |- (worker\suffix.west);
  }

  \node[overviewbox,draw=ThemeNavy!65,fill=ThemeNavy!5,
    minimum width=23pt,minimum height=16pt,right=12pt of stepsone]
    (deltaone) {\(\Delta_t^1\)};
  \node[overviewbox,draw=ThemeNavy!65,fill=ThemeNavy!5,
    minimum width=23pt,minimum height=16pt]
    at (deltaone |- workertwo) (deltatwo) {\(\Delta_t^2\)};
  \node[overviewbox,draw=ThemeNavy!65,fill=ThemeNavy!5,
    minimum width=23pt,minimum height=16pt]
    at (deltaone |- workerm) (deltam) {\(\Delta_t^M\)};
  \node[inner sep=0pt] at ($(deltatwo.south)!.5!(deltam.north)$) {\(\vdots\)};
  \begin{scope}[on background layer]
    \node[overviewbox,draw=ThemeNavy!45,fill=ThemeNavy!2,
      fit=(workerone)(deltaone)(workerm)(deltam),inner xsep=7pt,
      inner ysep=8pt] (workers) {};
  \end{scope}
  \node[overviewbox,draw=ThemeNavy!60,fill=ThemeNavy!8,
    font=\fontsize{8}{9}\selectfont\bfseries,
    minimum width=62pt,minimum height=14pt,inner sep=1pt]
    (workerheading) at (workers.north) {Large model: \(M\) workers};
  \draw[flow,dashed,draw=ScaleDistributed]
    (trainingrule.east) -- (87,-30) -- (87,54)
    -| (workerheading.north);
  \foreach \recipebox in {trainingrule,workerone,workertwo,workerm,innerone,innertwo,innerm} {
    \foreach \factorcolour/\offset in {ScaleWidth/0,ScaleDepth/9,ScaleBatch/18,ScaleTokens/27} {
      \fill[\factorcolour]
        ($(\recipebox.south)+(\offset pt,-0.2pt)+(-17.75pt,1pt)$)
        rectangle ++(8.5pt,1.6pt);
    }
  }
  \foreach \suffix in {one,two,m} {
    \draw[workerflow] (steps\suffix.east) -- (delta\suffix.west);
  }
  \coordinate (braceupper) at ($(deltaone.north east)+(11pt,0)$);
  \coordinate (bracelower) at ($(deltam.south east)+(11pt,0)$);
  \draw[decorate,decoration={brace,amplitude=3pt},line width=.65pt]
    (braceupper) -- (bracelower);
  \coordinate (averaged) at ($(braceupper)!.5!(bracelower)+(3pt,0)$);
  \node[overviewbox,draw=CalloutGold,fill=CalloutPeach,
    align=center,minimum width=47pt,minimum height=25pt,
    right=17pt of averaged] (outer)
    {\textbf{OuterOpt}\\[1pt]\((\gamma,\epsilon_{\mathrm{out}})\)};
  \draw[flow] (averaged) -- node[above,inner sep=1pt] {\(\Delta_t\)} (outer.west);
  \node[overviewbox,minimum width=23pt,minimum height=21pt,
    right=8pt of outer] (nexttheta) {\(\theta_{t+1}\)};
  \draw[flow] (outer.east) -- (nexttheta.west);
  \coordinate (averagingrule) at ($(outer.south)+(8pt,-5pt)$);
  \coordinate (linearrule) at ($(outer.south)+(8pt,-14pt)$);
  \coordinate (adamrule) at ($(outer.south)+(8pt,-23pt)$);
  \node[anchor=east,inner sep=0pt,font=\fontsize{7}{8}\selectfont]
    at ($(averagingrule)+(-15pt,0)$) {Averaging};
  \node[anchor=east,inner sep=0pt,font=\fontsize{7}{8}\selectfont]
    at ($(linearrule)+(-15pt,0)$) {Nesterov};
  \node[anchor=east,inner sep=0pt,font=\fontsize{7}{8}\selectfont]
    at ($(adamrule)+(-15pt,0)$) {Adam};
  \coordinate (averaginginherited) at ($(averagingrule)+(34pt,0)$);
  \coordinate (linearinherited) at ($(linearrule)+(34pt,0)$);
  \foreach \rulemark in {averagingrule,averaginginherited} {
    \foreach \factorcolour/\offset in {ScaleWidth/0,ScaleDepth/5.5,ScaleBatch/11,ScaleTokens/16.5} {
      \fill[\factorcolour] ($(\rulemark)+(\offset pt,-1.1pt)+(-10.5pt,0)$)
        rectangle ++(4.5pt,2.2pt);
    }
  }
  \foreach \rulemark in {linearrule,linearinherited,adamrule} {
    \fill[ScaleWidth] ($(\rulemark)+(-5pt,-1.1pt)$)
      rectangle ++(4.5pt,2.2pt);
    \fill[ScaleDepth] ($(\rulemark)+(.5pt,-1.1pt)$)
      rectangle ++(4.5pt,2.2pt);
  }
  \draw[flow] ($(averagingrule)+(12.5pt,0)$) -- ($(averaginginherited)+(-12.5pt,0)$);
  \draw[flow] ($(linearrule)+(7pt,0)$) -- ($(linearinherited)+(-7pt,0)$);
  \coordinate (adamfork) at ($(adamrule)+(11pt,0)$);
  \draw[line width=.65pt] ($(adamrule)+(7pt,0)$) -- (adamfork);
  \node[anchor=west,inner sep=0pt,font=\fontsize{7}{8}\selectfont]
    (gammarule) at ($(adamrule)+(23pt,4pt)$) {\(\gamma\)};
  \node[anchor=west,inner sep=0pt,font=\fontsize{7}{8}\selectfont]
    (epsilonrule) at ($(adamrule)+(23pt,-4pt)$) {\(\epsilon_{\mathrm{out}}\)};
  \draw[flow] (adamfork) |- ($(gammarule.west)+(-2pt,0)$);
  \draw[flow] (adamfork) |- ($(epsilonrule.west)+(-2pt,0)$);
\end{tikzpicture}
}
    \caption{\textbf{Applying \method.}
Tune the learning rate on a smaller model. Rescale initialization, forward
multipliers, and inner-optimizer settings for the target width, depth,
local batch size, and token budget, then apply the recipe on every
worker~\citep{yang2022tensorprograms,dey2025completep,mlodozeniec2025completedp}.
After $K$ local steps, synchronize worker updates and apply the outer optimizer.
Under averaging, the rules transfer across width, depth, and local batch
size.
Width and depth also transfer under \nesterov and, with corrections,
\adam. Batch and token-budget transfer with these outer optimizers may require further tuning.}
    \label{fig:distributed-scaling-overview}
\end{widefigure}

\section{\method: A Scaling Framework for Distributed Training}
\label{sec:new-distributed-training-scaling}

\method, as illustrated in \Cref{fig:distributed-scaling-overview},
enables zero-shot hyperparameter transfer in local-update training
(\cref{tab:distributed-scaling-recipe}).
Under suitable conditions, model averaging preserves width and depth
scaling~\citep{yang2022tensorprograms,dey2025completep} without further
corrections (\cref{sec:averaging-theory}).
More generally, linear outer optimizers preserve the same scaling,
while adaptive outer optimizers require learning-rate and epsilon
corrections (\cref{sec:outer-model-scaling}).
In local-update training, the global batch size can increase by
increasing each worker's local batch size or adding workers.
We show that, under averaging, the batch-scaling argument of
\citet{mlodozeniec2025completedp} extends to larger local batches,
but not to additional workers
(\cref{sec:worker-batch-duration-theory}).
Finally, we motivate local learning-rate transfer across synchronization
intervals under averaging, at fixed global batch and token budget
(\cref{sec:cadence-theory}).

\subsection{Width and depth transfer under averaging}
\label{sec:averaging-theory}
Let \(h_\ell(\theta)\) denote the activations at layer \(\ell\) on a fixed input.
For model averaging, we expand these features to first order under a parameter displacement \(\delta\theta\):
\[
 h_\ell(\theta_t+\delta\theta)
 =h_\ell(\theta_t)+J_{\ell,t}\delta\theta+r_\ell(\delta\theta).
\]
Here \(J_{\ell,t}\) is its Jacobian at the shared parameters
\(\theta_t\), and \(r_\ell\) is the
nonlinear remainder.
Worker \(m\)'s parameter displacement is \(-\Delta_t^m\), and the
averaged displacement is \(-\Delta_t\) (\cref{eq:aggregation}).
Multiplication by the shared Jacobian \(J_{\ell,t}\) is linear, giving
\(-J_{\ell,t}\Delta_t
=-\frac{1}{M}\sum_{m=1}^{M}J_{\ell,t}\Delta_t^m\). This shows that averaging parameter changes averages the first-order effects on features.

Existing width and depth parameterizations aim to keep feature changes
bounded and non-vanishing as model size grows~\citep{yang2022tensorprograms,dey2025completep}.
To extend this behavior to local-update training, we hold \(M\) and \(K\)
fixed and assume that the workers' accumulated first-order changes
\(-J_{\ell,t}\Delta_t^m\) remain bounded, with non-vanishing mean magnitude.
Our averaging argument uses two further conditions: the average must retain
a fixed positive fraction of that magnitude, and the nonlinear remainder
must stay below the averaged first-order change by a fixed relative margin.
Together, these assumptions give
\begin{equation}
 0<c_\ell
 \le
 \left\|h_\ell(\theta_t-\Delta_t)-h_\ell(\theta_t)\right\|
 \le C_\ell<\infty,
 \label{eq:server-average-feature-scale}
\end{equation}
where \(c_\ell,C_\ell\) are independent of width and depth. Thus, under
these assumptions, averaging preserves bounded, non-vanishing feature
changes without an additional width or depth scaling correction.
The full proof is in \cref{app:server-averaging-scaling}, with empirical coordinate checks in
\cref{app:toy-averaging,app:shakespeare-coordinates}.
\subsection{Width and depth transfer with outer optimizers}
\label{sec:outer-model-scaling}

Following~\Cref{sec:averaging-theory}, we examine whether other outer
optimizers preserve width and depth scaling.
Linear outer optimizers, such as \sgd \citep{robbins1951stochastic}
and \sgdm \citep{polyak1964some}, combine current and past
pseudo-gradients.
Here \(m_t\) denotes the outer optimizer's first moment, computed from the averaged pseudo-gradient \(\Delta_t\).
For outer \sgd, \cref{eq:outer-map} becomes
\(\theta_{t+1}=\theta_t-\gamma\Delta_t\), where
\(\gamma\) is the outer learning rate. With momentum,
\(m_t=\beta_{\mathrm{out}}m_{t-1}+\Delta_t\) and
\(\theta_{t+1}=\theta_t-\gamma m_t\). To test scaling,
suppose every pseudo-gradient is multiplied by \(s>0\).
Let \(T\) map pseudo-gradient history to the outer update.
With zero initial momentum and fixed hyperparameters, linearity gives
\(T(s\Delta_0,\ldots,s\Delta_t)
=s T(\Delta_0,\ldots,\Delta_t)\).
Thus, in the case of linear outer optimizers, scaling local updates by \(s\) scales the outer
update by \(s\), which requires no further width or depth correction.

Adaptive outer optimizers such as \adam~\citep{kingma2015adam} normalize this update. We denote the second moment of the outer optimizer by $v_t$.
Without weight decay, \cref{eq:outer-map} becomes
\(\theta_{t+1}=\theta_t-\gamma\widehat m_t/
(\sqrt{\widehat v_t}+\epsilon_{\mathrm{out}})\),
where \(\widehat m_t\) and
\(\widehat v_t\) are bias-corrected pseudo-gradient
moments \citep{kingma2015adam}. With zero-initialized moments
and fixed momentum coefficients, scaling every pseudo-gradient
by \(s\) scales these moments by \(s\) and \(s^2\), respectively. Unlike a linear outer optimizer, \adam leaves the update approximately
unchanged at fixed \(\gamma\) when epsilon is negligible before and
after scaling, or exactly unchanged if \(\epsilon_{\mathrm{out}}\)
also scales by \(s\).
To preserve the scaling per worker, \method introduces an additional multiplicative factor for \(\gamma\) and \(\epsilon_{\mathrm{out}}\), ensuring that the outer parameter update scales by \(s\)
(\cref{app:outer-scaling-contract}).

\subsection{Transfer across batch sizes}
\label{sec:worker-batch-duration-theory}
In local-update training, the global batch is computed as
\(B=Mb\), where \(M\) is the worker count and \(b\) the local batch size \citep{charles2025dilocoscaling}.
At fixed \(K\), \(B\) can increase through
either larger local batches $b$ at fixed worker count $M$ or more workers
at fixed local batch size.
Although prescriptions such as \completedp express batch scaling
in terms of \(B\) \citep{mlodozeniec2025completedp}, these
regimes affect optimization differently: larger local batches
change the gradients entering each worker's adaptive optimizer,
whereas additional workers increase the number of updates
combined by the outer optimizer.
We analyze both regimes at fixed token budget \(D\), so each
worker takes \(T_{\mathrm{inner}}=\frac{D}{Mb}\) steps.

\paragraph{Larger local batches, fixed workers.}
Increasing the local batch size has two effects: each worker receives
less noisy gradients, but takes fewer optimization steps at a
fixed token budget. We ask whether adjusting the local learning
rate can compensate for both effects. To examine this, we use
the \adam SDE approximation
\citep{malladi2022sde,mlodozeniec2025completedp} and consider two
quantities: the local noise parameter \(\eta\sigma_b\) and the
effective training horizon \(T_{\mathrm{inner}}\eta^2\),
where \(\sigma_b\) denotes the gradient noise scale. Increasing the local batch by a factor \(\kappa\), so that
\(b'=\kappa b\), reduces the gradient noise to
\(\sigma_{b'}=\sigma_b/\sqrt{\kappa}\) and the number of steps
to \(T'_{\mathrm{inner}}=T_{\mathrm{inner}}/\kappa\).
Preserving the local noise parameter requires
\(\eta'\sigma_{b'}=\eta\sigma_b\), which gives
\(\eta'=\sqrt{\kappa}\eta\).
This choice also preserves the effective training horizon:
\(T'_{\mathrm{inner}}(\eta')^2=T_{\mathrm{inner}}\eta^2\). These invariances recover the square-root learning-rate rule for \adam seen in prior work~\citep{malladi2022sde,mlodozeniec2025completedp}.

\begin{widetable}[t]
\centering
\small
\setlength{\tabcolsep}{7pt}
\renewcommand{\arraystretch}{1.75}

\caption{\textbf{\method and the choice of outer optimizer.}
Outer corrections for width/depth scaling and learning-rate transfer for
inner \adam at fixed tokens, varying one dimension at a time.
The \nesterov batch entry is an empirical starting prescription.
\(\eta,\gamma\) are inner and outer learning rates.
--- marks cases without an established rule.}
\label{tab:distributed-scaling-recipe}

\begin{tabular}{
    @{}>{\raggedright\arraybackslash}p{0.25\linewidth}
    >{\centering\arraybackslash}p{0.17\linewidth}
    >{\centering\arraybackslash}p{0.20\linewidth}
    >{\centering\arraybackslash}p{0.27\linewidth}@{}
}
\toprule
\shortstack{\textbf{Scaling dimension}\\~}
&
\shortstack{\textbf{Averaging}\\~}
&
\shortstack{\textbf{Linear momentum}\\
            \footnotesize (e.g., \nesterov)}
&
\shortstack{\textbf{Adaptive outer optimizer}\\
            \footnotesize (e.g., \adam)}
\\
\midrule

\shortstack{\textbf{Width / depth \;
\widthaxis{$N$}, \depthaxis{$L$}}\\~}
&
\shortstack{No outer\\correction}
&
\shortstack{No outer\\correction}
&
\shortstack{Scale $\gamma,\epsilon_{\mathrm{out}}$\\with width/depth}
\\

\addlinespace[3pt]

\textbf{Local batch}
\;
$\batchaxis{b}\rightarrow\kappa\batchaxis{b}$
&
$\eta\rightarrow\sqrt{\kappa}\eta$
&
\shortstack{Start $\eta'=\sqrt{\kappa}\eta$\\Tune $\gamma$ as needed}
&
---
\\

\addlinespace[3pt]

\textbf{Worker count}
\;
$\distaxis{M}\rightarrow\kappa\distaxis{M}$
&
$\eta$ unchanged
&
Tune $\eta$
&
---
\\

\addlinespace[3pt]

\textbf{Local steps}
\;
$\distaxis{K}$
&
$\eta$ unchanged
&
Tune $\eta,\gamma$
&
---
\\

\bottomrule
\end{tabular}
\end{widetable}

\paragraph{More workers, fixed local batch.}
Increasing the worker count divides the fixed token budget among
more workers.
With \(M'=\kappa M\), the step count therefore falls to
\(T'_{\mathrm{inner}}=T_{\mathrm{inner}}/\kappa\). However, each worker still
computes gradients from batches of size \(b\), so its gradient
noise scale remains \(\sigma_b\) at the same model parameters. Keeping \(\eta'=\eta\) preserves the local noise parameter,
\(\eta'\sigma_b=\eta\sigma_b\), but shortens the effective
training horizon to
\(T'_{\mathrm{inner}}(\eta')^2=T_{\mathrm{inner}}\eta^2/\kappa\).
Choosing \(\eta'=\sqrt{\kappa}\eta\) instead preserves the
horizon,
\(T'_{\mathrm{inner}}(\eta')^2=T_{\mathrm{inner}}\eta^2\),
but increases the local noise parameter to
\(\eta'\sigma_b=\sqrt{\kappa}\eta\sigma_b\). For \(\kappa\ne1\), there is no single learning-rate adjustment that preserves both
requirements, preventing a direct scaling rule from the SDE approximation of \adam
(\cref{app:rq2-sde-audit}).

\subsection{Transfer across local steps}
\label{sec:local-step-scaling}
\label{sec:cadence-theory}
At fixed global batch \(B\) and token budget \(D\), changing
\(K\) leaves the total number of local steps
\(T_{\mathrm{inner}}=D/B\) unchanged. Increasing \(K\)
therefore groups more steps into each communication round
while reducing the number of rounds,
\(R_{\mathrm{comm}}=T_{\mathrm{inner}}/K\). Under averaging with constant
learning rate \(\eta\), the total step-size sum is
\begin{equation}
 R_{\mathrm{comm}}K\eta=T_{\mathrm{inner}}\eta.
 \label{eq:fixed-budget-cadence}
\end{equation}
Under averaging, we retain the reference learning rate as \(K\) changes,
preserving this total step-size sum. Longer intervals can increase divergence
between workers' parameters and \adam moments, which is not addressed by this argument. We leave characterizing learning-rate transfer
when \(K\) greatly exceeds the inner momentum half-life to future work. With outer momentum, changing \(K\) alters its update frequency and
memory in local steps. These changes motivate joint inner--outer
learning-rate tuning~\citep{khaled2025outer}, as explored in \cref{sec:rq3}.

\paragraph{Applying \method.}
\label{sec:transfer-recipe}
As shown in~\Cref{fig:distributed-scaling-overview}, \method combines width and depth scaling on each worker with
local batch-size and token-budget prescriptions under model
averaging. Starting from a tuned reference configuration, we apply
the prescribed initialization and forward multipliers on every worker. We then multiply
each local optimizer's reference hyperparameters by the model and training
factors~\citep{dey2025completep,mlodozeniec2025completedp}.
For hidden matrices $\mathrm{h}$:
\begin{equation}
 \eta_{\mathrm{h}}=\eta_{\mathrm{h},0}
 \times\widthaxis{\frac{1}{m_N}}
 \times\depthaxis{m_L^{\alpha-1}}
 \times\sqrt{\frac{\batchaxis{m_B}}{\dataaxis{m_D}}},
 \label{eq:composed-inner-lr}
\end{equation}
where \(m_B=b/b_0\). Each worker uses these settings in
\(\operatorname{InnerOpt}\) (\cref{eq:inner-map}). At synchronization,
averaging applies the resulting pseudo-gradient \(\Delta_t\)
with outer learning rate \(\gamma=1\).
For other linear outer optimizers, no further width or depth correction
is needed. Outer \adam\ applies
the corresponding model factors to its learning rate and epsilon
in \(\operatorname{OuterOpt}\) (\cref{eq:outer-map}).

\singlecolumnfloatbarrier

\section{Experiments}
\raggedbottom
Building on the analysis in \cref{sec:new-distributed-training-scaling},
we investigate the following research questions:
\begin{enumerate}[label=\textbf{RQ\arabic*}, leftmargin=*,itemsep=2pt,
    before={\interlinepenalty=10000\relax}]
    \item Does averaging preserve learning-rate transfer across width and depth?
    \item How do local batches, worker counts, and local steps affect transfer under averaging?
    \item How do outer optimizers affect transfer across model and training configurations?
    \item How does \method scale with model size and training budget?
\end{enumerate}
\subsection{Setup}
\label{sec:experimental-setup}
\paragraph{Models and data.}
We train Transformer models~\citep{vaswani2017attention} using a pre-norm, decoder-only
architecture on a SmolLM-based mixture of
web, educational, code and mathematical text~\citep{allal2025smollm2},
using three model scales: \smallmodel (\(29.2\)M),
\mediummodel (\(162.3\)M), and \largemodel (\(822.9\)M).
\smallmodel acts as the reference architecture for our width and depth
sweeps. We evaluate transfer across training configurations at
\smallmodel and \mediummodel scales, and test scaling
on \largemodel over a longer horizon.
See \cref{app:experimental-setup} for architecture and data details.

\paragraph{Optimization and transfer.}
We use \adamw~\citep{loshchilov2017decoupled} with zero weight decay
(equivalent to \adam) as \texttt{InnerOpt} and synchronize parameters
and both moments at the same cadence, following \desloc~\citep{iacob2025desloc}.
For \texttt{OuterOpt}, we use \sgd with unit learning rate
(model averaging), \nesterov, or \adam.
Unless otherwise stated, we set \(M=4\) and \(K=32\).
For \method, we apply the zero-shot transfer rules to initialization,
forward multipliers, and inner-optimizer settings on every worker, with outer corrections from
\cref{sec:new-distributed-training-scaling}. For the batch and synchronization sweeps, we fix the token budget,
the inner \adam moment coefficients, and epsilon.
We also fix the global batch when varying the synchronization interval.
Full settings and sweep protocols are given in
\cref{app:experiment-protocols}.

\paragraph{Comparisons.}
We compare \method with standard parameterization (\standardp),
which omits the width and depth scaling factors, using matched
architectures, batch sizes, and training data.
We first sweep learning rates to assess transfer across
model and training configurations. We then compare model quality
at matched compute, using each architecture
and parameterization's best-tested inner learning rate at each compute budget.
Finally, at \largemodel scale, we compare settings transferred from
\smallmodel against \dilocofit, which uses \standardp with the learning-rate
prescription of \citet{charles2025dilocoscaling}, without target-scale retuning.

\paragraph{Metrics.}
We measure model quality using validation loss on a fixed holdout
from the training mixture.
To assess whether improvements extend to downstream knowledge
and reasoning tasks, we evaluate \largemodel accuracy on ARC-Easy and
ARC-Challenge~\citep{clark2018arc},
HellaSwag~\citep{zellers2019hellaswag},
PIQA~\citep{bisk2019piqa}, and MMLU~\citep{hendrycks2020mmlu}.
Additional experiments and ablations are provided in
\cref{app:additional-results}.

\subsection{\method transfers across width and depth under averaging (RQ1)}
\label{sec:rq1}
\label{sec:model-transfer-averaging}
Supporting our derivation in \Cref{sec:averaging-theory},
\Cref{fig:rq1-width} shows that \method preserves learning-rate
transfer as width increases while achieving lower validation loss
than \standardp, whose preferred learning rate shifts with width.
This demonstrates the width-transfer behavior of
\mup~\citep{yang2022tensorprograms} in the local-update setting.
Across depth, \method also preserves learning-rate transfer,
consistent with \citet{dey2025completep}. We observe that \standardp also shows approximate
learning-rate alignment, but \method achieves lower validation
loss (\cref{fig:rq1-depth}).
We expect this alignment to break down at greater depths,
consistent with the depth-transfer failures reported by
\citet{dey2025completep}. Following \citet{mlodozeniec2025completedp}, we also ablate
Q/K normalization, finding that \standardp remains more
sensitive at high learning rates
(\cref{fig:rq1-qk-off-depth}).

\begin{widefigure}[!htbp]
 \centering
 \rqsplitpanels{rq1-width-averaging}{fig:rq1-width}{width}
 \caption{\textbf{\method preserves width transfer under averaging.}
 At \(M=4,K=32\), colors denote width relative to \(256\). Points show final validation loss across inner learning rates.}
 \label{fig:rq1-width}
\end{widefigure}

\begingroup\tcbset{before skip=4pt,after skip=4pt}
\takeawaybox[Width and depth transfer for averaging.]{
Under averaging, \method preserves empirical width and depth learning-rate transfer
while improving model quality relative to \standardp.
}
\endgroup

\singlecolumnfloatbarrier

\vspace{-4pt}
\subsection{Batch-size and local-step transfer under averaging (RQ2)}
\label{sec:rq2}
\label{sec:worker-batch-evaluation}
\label{sec:cadence-evaluation}
\begin{widefigure}[!htb]
 \centering
 \renewcommand{\rqcadenceheading}{Inner learning rate transfer}
 \rqregimepanels{rq2-averaging-medium-regimes}{fig:rq2-worker-count-main}{fig:rq2-local-batch-main}{fig:rq3-cadence-summary}
\caption{\textbf{Larger local batches favor square-root scaling, while
workers and local steps favor unchanged rates.}
Averaging with inner \adam on \mediummodel at fixed tokens.
(a) Worker-count sweeps with reference guide \(\eta_0\).
(b) Local-batch sweeps with \adam SDE guides
\(\eta_0\sqrt{b/b_0}\). The dotted path joins best tested rates.
(c) Half, reference, and double rate across synchronization
intervals at fixed global batch. Excess loss is relative to
each interval's best tested rate.
Reference: \(M_0=4\), \(b_0=64\) sequences, \(K=32\),
\(\eta_0=.016\).}
 \label{fig:rq2-regimes}
 \label{fig:rq2-batch-worker-summary}
\end{widefigure}

Supporting the \adam SDE argument in
\cref{sec:worker-batch-duration-theory}, we find that the best tested learning rate scales with the square root of local batch size,
achieving lower validation loss than retaining the reference
rate (\cref{fig:rq2-local-batch-main}).
By contrast, increasing the worker count at fixed local batch
size favors retaining the reference learning rate
(\cref{fig:rq2-worker-count-main}).
This reinforces that, although both changes increase global batch size,
additional workers increase the number of updates averaged
without reducing each worker's local gradient noise.
At fixed global batch and token budget, the local
rate transfers unchanged across synchronization
intervals (\cref{fig:rq3-cadence-summary}), supporting
\cref{sec:cadence-theory}.
See \cref{fig:rq2-small-regimes} for \smallmodel results.

\begingroup\tcbset{before skip=4pt,after skip=4pt}
\takeawaybox[Learning-rate transfer under averaging.]{
Larger local batches favor the square-root rule from the
\adam SDE, whereas additional workers favor retaining the reference rate.
Changing the synchronization interval requires no learning-rate
adjustment at fixed global batch and token budget.
}
\endgroup

\subsection{Outer optimizers preserve model-size transfer (RQ3)}
\label{sec:rq3}
\label{sec:outer-transfer}
{\interlinepenalty=10000
Following \Cref{sec:rq2}, we turn to examine \method's generalization beyond averaging. Without additional corrections to outer \nesterov (\cref{sec:outer-model-scaling}), \method preserves width and depth transfer and achieves lower validation loss than \standardp (\cref{fig:rq3-width,fig:rq1-depth}). Like \citet{charles2025dilocoscaling}, we also find that the outer learning rate remains effective across model sizes (\cref{app:outer-protocol}). Notably, we find that \method's learning-rate and epsilon corrections (\cref{sec:outer-model-scaling}) enable stable width and depth transfer with outer \adam, achieving lower loss than \standardp. This shows that \adam can be an effective outer optimizer, which \citet{douillard2023diloco} found less effective than \nesterov. Next, we examine batch scaling with \nesterov as the outer optimizer. The \adam SDE's square-root rule~\citep{malladi2022sde,mlodozeniec2025completedp}
is an effective starting point for larger local batches
(\cref{fig:rq3-nesterov-regimes,fig:rq3-small-regimes}). Outer-rate tuning
can narrow the gap to an unchanged inner rate (\cref{app:joint-retuning}).
Joint tuning also reduces loss at some synchronization intervals,
consistent with \citet{khaled2025outer}.
Joint inner--outer transfer rules are left for future work.\par}

\begin{widefigure}[!htb]
 \centering
 \begin{minipage}{.88\linewidth}\centering
 \rqmatrixpanels{width}{fig:rq3-width}{0}
 \end{minipage}
 \caption{\textbf{Width transfer extends to momentum and adaptive outer updates.}
 Rows show \nesterov and \adam as \texttt{OuterOpt} at \(M=4,K=32\).
 Colors denote width relative to \(256\). Points show final validation loss across inner learning rates.}
 \label{fig:rq3-width}
\end{widefigure}

\nopagebreak[4]
\begingroup\tcbset{before skip=2pt,after skip=2pt}
\takeawaybox[Outer momentum couples the learning rates.]{
Width and depth transfer extend to \nesterov and \adam.
With \nesterov, start with square-root local-batch scaling and tune
\(\gamma\) as needed. Synchronization changes can benefit from
joint tuning~\citep{khaled2025outer}.
}
\endgroup

\begin{widefigure}[!htb]
 \centering
 \begin{minipage}{.90\linewidth}\centering
 \rqregimepanels{rq3-nesterov-medium-regimes}{fig:rq3-nesterov-workers}{fig:rq3-nesterov-local-batch}{fig:rq3-nesterov-local-steps}
 \end{minipage}
\caption{\textbf{\nesterov couples inner and outer learning rates.}
\nesterov with inner \adam on \mediummodel at fixed tokens.
(a,b) Worker-count and local-batch sweeps at fixed outer rate
\(\gamma_0\), with guides at \(\eta_0\) and
\(\eta_0\sqrt{b/b_0}\), respectively.
The dotted path joins best tested rates in (b).
(c) Half, reference, and double outer rate across synchronization
intervals at fixed inner rate \(\eta_0\) and global batch.
Excess loss is relative to each interval's best tested outer rate.
Reference: \(M_0=4\), \(b_0=64\), \(K=32\),
\(\eta_0=.016\), \(\gamma_0=.2\).
Corresponding \smallmodel results are in 
\cref{fig:rq3-small-regimes}.}
 \label{fig:rq3-nesterov-regimes}

\end{widefigure}

\singlecolumnfloatbarrier
\suppressfloats[t]
\subsection{Improved performance across model sizes and training budgets (RQ4)}
\label{sec:rq4}
\begin{widetable}[tb]
 \centering\small
 \setlength{\tabcolsep}{3.5pt}
 \caption{\textbf{Small-tuned \method outperforms downstream.}
 Accuracy (\%) at \(823\)M parameters and \(17.18\)B tokens.
 Evaluation is zero-shot except for five-shot MMLU. Bold marks maxima.}
 \label{tab:rq4-downstream}
 \begin{tabular}{@{}llrrrrrr@{}}
 \toprule
 Outer update & Recipe & ARC-E & ARC-C & HellaSwag & PIQA & MMLU & Mean\\
 \midrule
 Averaging & \dilocofit & \(49.92\) & \(28.24\) & \(34.57\) & \(66.32\) & \(25.99\) & \(41.01\)\\
 & \method{} (Reference) & \(55.60\) & \(\mathbf{31.06}\) & \(41.36\) & \(68.17\) & \(25.79\) & \(44.39\)\\
 & \method{} (Small-tuned) & \(56.82\) & \(29.69\) & \(\mathbf{43.08}\) & \(\mathbf{69.48}\) & \(25.31\) & \(44.88\)\\
 \midrule
 \nesterov & \dilocofit & \(50.51\) & \(26.96\) & \(34.95\) & \(66.65\) & \(26.42\) & \(41.10\)\\
 & \method{} (Reference) & \(52.10\) & \(28.58\) & \(38.12\) & \(67.19\) & \(25.74\) & \(42.35\)\\
 & \method{} (Small-tuned) & \(\mathbf{58.96}\) & \(30.20\) & \(41.92\) & \(68.23\) & \(\mathbf{26.57}\) & \(\mathbf{45.18}\)\\
 \bottomrule
 \end{tabular}
\end{widetable}

We first compare \method and \standardp on validation loss at matched
FLOP budgets, accounting for model size and training
duration~\citep{hoffmann2022compute}.
We use each architecture and parameterization's best-tested
inner learning rate at each compute budget.
We focus on averaging, which requires no
outer-rate tuning.
In~\Cref{fig:rq4-scaling-averaging}, we find that \method
consistently achieves lower loss than \standardp across FLOP
budgets. The fitted scaling laws suggest that \method's
advantage grows with compute (\cref{app:compute-frontier}),
with similar trends under \nesterov
(\cref{fig:rq4-scaling-nesterov}).

We then compare hyperparameter prescriptions at \largemodel
scale without target-scale tuning, using a shared training protocol.
Our reference is the four-worker
learning-rate law of \citet{charles2025dilocoscaling},
\(\eta_{\mathrm C}(P)=4512.99P^{-0.789}\), where \(P\) refers
to the parameter count.
The \dilocofit baseline evaluates this law directly at
\largemodel scale under \standardp, whereas \method{} (Reference)
evaluates it at \smallmodel and transfers using our scaling rules.
We compare these recipes under both averaging and \nesterov
as the outer optimizer, retaining the published outer rate
\(\gamma=.6\) for \nesterov~\citep{charles2025dilocoscaling}.
However, we note that fitted hyperparameter laws may require recalibration across training setups.
To account for this, we tune \diloco for our setup at \smallmodel scale,
following \citet{therien2025muloco}, and transfer these settings with
\method{} (Small-tuned).
This recipe uses the settings selected in \cref{sec:rq1,sec:rq3},
including outer rate \(\gamma=.1\) for \nesterov
(\cref{app:rq4-large-protocol}).
As shown in \cref{fig:rq4-large-validation}, both \method
recipes achieve lower validation loss than \dilocofit under
both outer optimizers, with further improvements from
Small-tuned settings.
To assess whether these gains extend beyond validation loss,
we also evaluate accuracy on downstream knowledge and reasoning tasks.
At \(823\)M parameters, \method{} (Small-tuned) with \nesterov
reduces validation loss by \(8.1\%\) and improves mean downstream
accuracy by \(10\%\) relative to \dilocofit
(\cref{tab:rq4-downstream}). These results show that \method improves model
quality at matched compute and delivers large-model gains
through small-scale tuning, without fitting a new learning-rate
law or retuning at the target scale.

\nopagebreak[4]
\takeawaybox[Small-model tuning improves large-model training.]{
\method improves model quality at matched compute and
large-model performance through small-model tuning,
without target-scale retuning.
}
\begin{widefigure}[!htb]
 \centering
 \begin{subfigure}[t]{.5247\linewidth}\centering
  \captionsetup{position=top,skip=3pt}\caption{Width and depth scaling}\label{fig:rq4-compute-trajectories}
  \includegraphics[width=\linewidth]{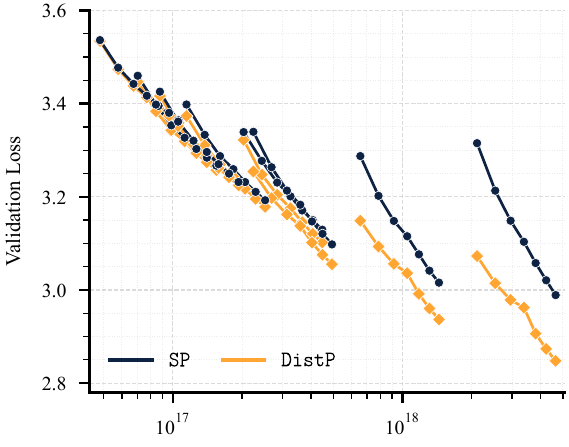}
 \end{subfigure}\hfill
 \begin{subfigure}[t]{.4553\linewidth}\centering
  \captionsetup{position=top,skip=3pt}\caption{Loss--compute fit}\label{fig:rq4-compute-frontier}
  \includegraphics[width=\linewidth]{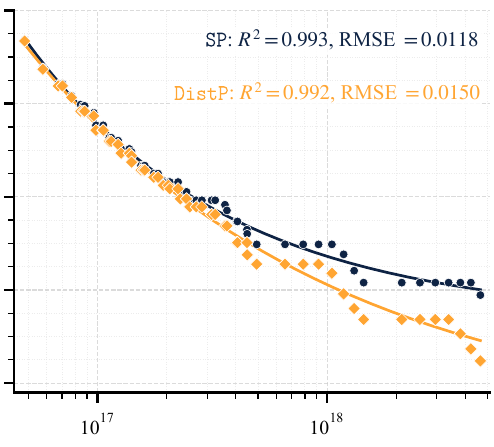}
 \end{subfigure}\par\vspace{1pt}FLOPs
 \caption{\textbf{\method improves loss at matched compute under averaging.}
 (a) Validation loss using each architecture and parameterization's
 best-tested inner learning rate at each measured compute budget.
 (b) Best observed loss at or below each FLOP budget, fitted with the empirical
 offset power law \(L(C)=E+A(C/10^{18})^{-\alpha}\). Annotations give in-sample fit quality.
 }
 \label{fig:rq4-scaling-averaging}
\end{widefigure}

\begin{widefigure}[!htb]
 \centering
 \includegraphics[width=.93\linewidth]{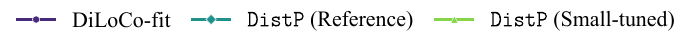}\par\nointerlineskip
 \begin{subfigure}[t]{.5247\linewidth}\centering
  \captionsetup{position=top,skip=3pt}\caption{Averaging}\label{fig:rq4-large-averaging}
  \includegraphics[width=\linewidth]{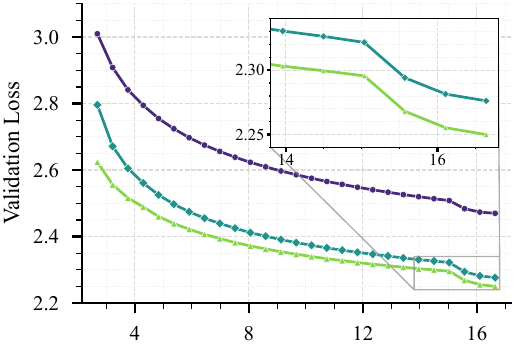}
 \end{subfigure}\hfill
 \begin{subfigure}[t]{.4553\linewidth}\centering
  \captionsetup{position=top,skip=3pt}\caption{\nesterov}\label{fig:rq4-large-nesterov}
  \includegraphics[width=\linewidth]{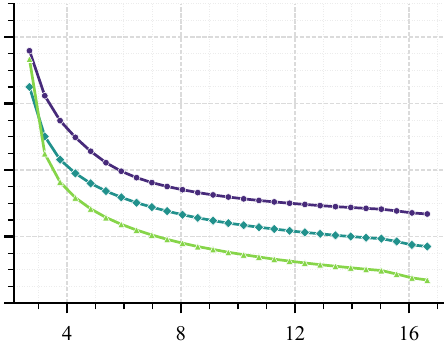}
 \end{subfigure}\par\vspace{1pt}Training tokens (billions)
 \caption{\textbf{Transferred recipes improve Large-model training.}
 Validation loss at \(823\)M parameters. \dilocofit applies the published
 inner-rate law~\citep{charles2025dilocoscaling} at \largemodel under standard
 parameterization. \method instead transfers Reference or Small-tuned
 settings from \smallmodel.}
 \label{fig:rq4-large-validation}
\end{widefigure}
\singlecolumnfloatbarrier

\section{Related Work}
\paragraph{Hyperparameter transfer.}
Existing parameterizations reduce retuning by specifying how
training settings change with scale.
\mup~\citep{yang2022tensorprograms} provides width transfer,
\completep~\citep{dey2025completep} extends it to depth, and
\completedp~\citep{mlodozeniec2025completedp} incorporates
batch size and token budget.
\citet{charles2025dilocoscaling} address hyperparameter transfer
in \diloco by fitting learning rates as a function of model
size, validating their prescriptions at \(4\)B and \(10\)B parameters.
\method addresses the same tuning problem by extending
existing zero-shot hyperparameter transfer rules to local-update training,
enabling transfer from a small reference configuration
without fitting a new model-size law.

\paragraph{Communication-efficient training.}
\localsgd~\citep{stich2019local} and \fedavg~\citep{mcmahan2017fedavg} reduce communication by allowing workers
to optimize locally before averaging
parameters, an approach
also adopted for language-model training in \diloco~\citep{douillard2023diloco} and
\photon~\citep{sani2025photon}.
\slowmo~\citep{wang2020slowmo} and
\fedopt~\citep{reddi2021adaptive} incorporate momentum or
adaptive outer updates, making the interaction between inner
and outer optimization central to hyperparameter selection.
For quadratic objectives, \citet{charles2020outsized} show
that the inner learning rate changes the effective objective,
while \citet{khaled2025outer} establish how the outer rate
balances optimization error and gradient noise.
Beyond learning rates, inner preconditioning also affects
replica scaling~\citep{belilovsky2026replica}.
Building on these findings, \method extends hyperparameter
transfer to local-update training by accounting for both
inner and outer optimization.
We provide an extended discussion of related work in \cref{app:related-work}.

\section{Discussion and Conclusion}
In this paper, we introduce \method, which extends zero-shot
hyperparameter transfer to local-update training by applying
existing transfer rules on every worker and accounting for
the outer optimizer.
Our experiments show improved model quality, relative to \standardp, at matched compute
and effective transfer from small to large models.
This provides a principled way to scale local-update training from a tuned reference
configuration without fitting a new learning-rate law across
model sizes.

Although width and depth transfer extend to \nesterov and,
with corrections, outer \adam, batch and synchronization
transfer with \nesterov can still benefit from further tuning.
For larger local batches, square-root inner learning-rate scaling provides
an effective starting point.
Averaging, however, achieves competitive quality with simpler
transfer and no outer learning rate to tune, making it a more effective
and scalable choice for local-update training.
We invite further theoretical work on joint transfer rules
for inner--outer optimizer combinations to avoid
repeated retuning.

\FloatBarrier
\endgroup
\section*{Acknowledgements}
We gratefully acknowledge core support from the Science of Fundamental AI Research (SOFAIR) Lab, funded by the Engineering and Physical Sciences Research Council (EPSRC), part of UK Research and Innovation (UKRI), and from the Royal Academy of Engineering through its Chair in Emerging Technologies supporting DANTE. This research also received support from the European Research Council through the REDIAL project; SPRIND through the Composite Learning Challenge; the Foresight Institute through an AI Safety Grant; Google through a Google Academic Research Award; IMEC; and the Ministry of Education of Romania through the Credit and Scholarship Agency. This research was partially supported by the DFG Cluster of Excellence MATH+ (EXC-2046/2,
project id 390685689) funded by the Deutsche Forschungsgemeinschaft (DFG) as well as by the
German Federal Ministry of Research, Technology and Space (research campus Modal, fund number 05M14ZAM, 05M20ZBM). Christophe Roux was supported by the French government, managed by the Agence Nationale de la Recherche, under the France 2030 program (PR[AI]RIE-PSAI) and the Investissements d'avenir program (PRAIRIE 3IA Institute) [ANR-23-IACL-0008].

\begingroup
\makeatletter
\ifpaper@twocolumn\else
  \setlength{\bibsep}{6pt plus 1pt minus 1pt}
\fi
\makeatother
\bibliographystyle{\paperbibstyle}
\bibliography{references}

\begin{thebibliography}{48}
\providecommand{\natexlab}[1]{#1}
\providecommand{\url}[1]{\texttt{#1}}
\expandafter\ifx\csname urlstyle\endcsname\relax
  \providecommand{\doi}[1]{doi: #1}\else
  \providecommand{\doi}{doi: \begingroup \urlstyle{rm}\Url}\fi

\bibitem[Narayanan et~al.(2021)Narayanan, Shoeybi, Casper, LeGresley, Patwary,
  Korthikanti, Vainbrand, Kashinkunti, Bernauer, Catanzaro, Phanishayee, and
  Zaharia]{narayanan2021efficient}
Deepak Narayanan, Mohammad Shoeybi, Jared Casper, Patrick LeGresley, Mostofa
  Patwary, Vijay Korthikanti, Dmitri Vainbrand, Prethvi Kashinkunti, Julie
  Bernauer, Bryan Catanzaro, Amar Phanishayee, and Matei Zaharia.
\newblock Efficient large-scale language model training on {GPU} clusters using
  {Megatron-LM}.
\newblock In \emph{Proceedings of the International Conference for High
  Performance Computing, Networking, Storage and Analysis}, 2021.

\bibitem[Douillard et~al.(2024)Douillard, Feng, Rusu, Chhaparia, Donchev,
  Kuncoro, Ranzato, Szlam, and Shen]{douillard2023diloco}
Arthur Douillard, Qixuan Feng, Andrei~A. Rusu, Rachita Chhaparia, Yani Donchev,
  Adhiguna Kuncoro, Marc'Aurelio Ranzato, Arthur Szlam, and Jiajun Shen.
\newblock {DiLoCo}: Distributed low-communication training of language models.
\newblock \emph{arXiv preprint arXiv:2311.08105}, 2024.

\bibitem[Iacob et~al.(2026{\natexlab{a}})Iacob, Jovanovic, Safaryan, Kurmanji,
  Sani, Horv{\'a}th, Shen, Qiu, and Lane]{iacob2025mtdao}
Alex Iacob, Andrej Jovanovic, Mher Safaryan, Meghdad Kurmanji, Lorenzo Sani,
  Samuel Horv{\'a}th, William~F. Shen, Xinchi Qiu, and Nicholas~D. Lane.
\newblock {MT}-{DAO}: Multi-timescale distributed adaptive optimizers with
  local updates.
\newblock In \emph{International Conference on Learning Representations
  (ICLR)}, 2026{\natexlab{a}}.

\bibitem[Jovanovic et~al.(2026)Jovanovic, Iacob, Safaryan, Modoranu, Sani,
  Shen, Qiu, Alistarh, and Lane]{jovanovic2026lordo}
Andrej Jovanovic, Alex Iacob, Mher Safaryan, Ionut-Vlad Modoranu, Lorenzo Sani,
  William~F. Shen, Xinchi Qiu, Dan Alistarh, and Nicholas~D. Lane.
\newblock Lo{RDO}: Distributed low-rank optimization with infrequent
  communication.
\newblock In \emph{International Conference on Machine Learning (ICML)}, 2026.

\bibitem[McMahan et~al.(2017)McMahan, Moore, Ramage, Hampson, and
  Arcas]{mcmahan2017fedavg}
Brendan McMahan, Eider Moore, Daniel Ramage, Seth Hampson, and Blaise Aguera~y
  Arcas.
\newblock {Communication-Efficient Learning of Deep Networks from Decentralized
  Data}.
\newblock In \emph{International Conference on Artificial Intelligence and
  Statistics (AISTATS)}, 2017.

\bibitem[Stich(2019)]{stich2019local}
Sebastian~U. Stich.
\newblock Local {SGD} converges fast and communicates little.
\newblock In \emph{International Conference on Learning Representations
  (ICLR)}, 2019.

\bibitem[Wang et~al.(2020)Wang, Tantia, Ballas, and Rabbat]{wang2020slowmo}
Jianyu Wang, Vinayak Tantia, Nicolas Ballas, and Michael Rabbat.
\newblock {SlowMo}: Improving communication-efficient distributed {SGD} with
  slow momentum.
\newblock In \emph{International Conference on Learning Representations
  (ICLR)}, 2020.

\bibitem[Reddi et~al.(2021)Reddi, Charles, Zaheer, Garrett, Rush,
  Kone{\v{c}}n{\'y}, Kumar, and McMahan]{reddi2021adaptive}
Sashank~J. Reddi, Zachary Charles, Manzil Zaheer, Zachary Garrett, Keith Rush,
  Jakub Kone{\v{c}}n{\'y}, Sanjiv Kumar, and Hugh~Brendan McMahan.
\newblock Adaptive federated optimization.
\newblock In \emph{International Conference on Learning Representations
  (ICLR)}, 2021.

\bibitem[Charles et~al.(2025)Charles, Teston, Dery, Rush, Fallen, Garrett,
  Szlam, and Douillard]{charles2025dilocoscaling}
Zachary Charles, Gabriel Teston, Lucio~M. Dery, J~Keith Rush, Nova Fallen,
  Zachary Garrett, Arthur Szlam, and Arthur Douillard.
\newblock Communication-efficient language model training scales reliably and
  robustly: Scaling laws for {DiLoCo}.
\newblock In \emph{Conference on Neural Information Processing Systems
  (NeurIPS)}, 2025.

\bibitem[Th{\'e}rien et~al.(2026)Th{\'e}rien, Huang, Defazio, Rish, and
  Belilovsky]{therien2025muloco}
Benjamin Th{\'e}rien, Xiaolong Huang, Aaron Defazio, Irina Rish, and Eugene
  Belilovsky.
\newblock {MuLoCo}: Muon is a practical inner optimizer for {DiLoCo}.
\newblock In \emph{International Conference on Machine Learning (ICML)}, 2026.

\bibitem[Yang et~al.(2021)Yang, Hu, Babuschkin, Sidor, Liu, Farhi, Ryder,
  Pachocki, Chen, and Gao]{yang2022tensorprograms}
Greg Yang, Edward~J Hu, Igor Babuschkin, Szymon Sidor, Xiaodong Liu, David
  Farhi, Nick Ryder, Jakub Pachocki, Weizhu Chen, and Jianfeng Gao.
\newblock Tuning large neural networks via zero-shot hyperparameter transfer.
\newblock In \emph{Conference on Neural Information Processing Systems
  (NeurIPS)}, 2021.

\bibitem[Dey et~al.(2025)Dey, Zhang, Noci, Li, Bordelon, Bergsma, Pehlevan,
  Hanin, and Hestness]{dey2025completep}
Nolan~Simran Dey, Bin~Claire Zhang, Lorenzo Noci, Mufan Li, Blake Bordelon,
  Shane Bergsma, Cengiz Pehlevan, Boris Hanin, and Joel Hestness.
\newblock Don't be lazy: {CompleteP} enables compute-efficient deep
  transformers.
\newblock In \emph{Conference on Neural Information Processing Systems
  (NeurIPS)}, 2025.

\bibitem[Mlodozeniec et~al.(2026)Mlodozeniec, Ablin, B{\'e}thune, Busbridge,
  Klein, Ramapuram, and Cuturi]{mlodozeniec2025completedp}
Bruno~Kacper Mlodozeniec, Pierre Ablin, Louis B{\'e}thune, Dan Busbridge,
  Michal Klein, Jason Ramapuram, and Marco Cuturi.
\newblock Completed hyperparameter transfer across modules, width, depth, batch
  and duration.
\newblock In \emph{International Conference on Learning Representations
  (ICLR)}, 2026.

\bibitem[Khaled et~al.(2025)Khaled, Kale, Douillard, Jin, Fergus, and
  Zaheer]{khaled2025outer}
Ahmed Khaled, Satyen Kale, Arthur Douillard, Chi Jin, Rob Fergus, and Manzil
  Zaheer.
\newblock Understanding outer optimizers in local {SGD}: Learning rates,
  momentum, and acceleration.
\newblock In \emph{Conference on Neural Information Processing Systems
  (NeurIPS)}, 2025.

\bibitem[Iacob et~al.(2026{\natexlab{b}})Iacob, Sani, Safaryan, Giampouras,
  Horv{\'a}th, Jovanovic, Kurmanji, Aleksandrov, Shen, Qiu, and
  Lane]{iacob2025desloc}
Alex Iacob, Lorenzo Sani, Mher Safaryan, Paris Giampouras, Samuel Horv{\'a}th,
  Andrej Jovanovic, Meghdad Kurmanji, Preslav Aleksandrov, William~F. Shen,
  Xinchi Qiu, and Nicholas~D. Lane.
\newblock {DES}-{LOC}: Desynced low communication adaptive optimizers for
  foundation models.
\newblock In \emph{International Conference on Learning Representations
  (ICLR)}, 2026{\natexlab{b}}.

\bibitem[Cheng and Glasgow(2025)]{cheng2025convergence}
Ziheng Cheng and Margalit Glasgow.
\newblock Convergence of distributed adaptive optimization with local updates.
\newblock In \emph{International Conference on Learning Representations
  (ICLR)}, 2025.

\bibitem[Robbins and Monro(1951)]{robbins1951stochastic}
Herbert~E. Robbins and Sutton Monro.
\newblock A stochastic approximation method.
\newblock \emph{Annals of Mathematical Statistics}, 22:\penalty0 400--407,
  1951.

\bibitem[Polyak(1964)]{polyak1964some}
Boris Polyak.
\newblock Some methods of speeding up the convergence of iteration methods.
\newblock \emph{{USSR} Computational Mathematics and Mathematical Physics},
  4:\penalty0 1--17, 1964.

\bibitem[Kingma and Ba(2015)]{kingma2015adam}
Diederik~P. Kingma and Jimmy Ba.
\newblock Adam: A method for stochastic optimization.
\newblock In \emph{International Conference on Learning Representations
  (ICLR)}, 2015.

\bibitem[Malladi et~al.(2022)Malladi, Lyu, Panigrahi, and
  Arora]{malladi2022sde}
Sadhika Malladi, Kaifeng Lyu, Abhishek Panigrahi, and Sanjeev Arora.
\newblock On the {SDE}s and scaling rules for adaptive gradient algorithms.
\newblock In \emph{Conference on Neural Information Processing Systems
  (NeurIPS)}, 2022.

\bibitem[Vaswani et~al.(2017)Vaswani, Shazeer, Parmar, Uszkoreit, Jones, Gomez,
  Kaiser, and Polosukhin]{vaswani2017attention}
Ashish Vaswani, Noam Shazeer, Niki Parmar, Jakob Uszkoreit, Llion Jones,
  Aidan~N. Gomez, {\L}ukasz Kaiser, and Illia Polosukhin.
\newblock Attention is all you need.
\newblock In \emph{Conference on Neural Information Processing Systems
  (NeurIPS)}, 2017.

\bibitem[Allal et~al.(2025)Allal, Lozhkov, Bakouch, Blazquez, Penedo, Tunstall,
  Marafioti, Lajar{\'\i}n, Kydl{\'\i}{\v{c}}ek, Srivastav, Lochner, Fahlgren,
  NGUYEN, Burtenshaw, Fourrier, Zhao, Larcher, Morlon, Zakka, Raffel, Werra,
  and Wolf]{allal2025smollm2}
Loubna~Ben Allal, Anton Lozhkov, Elie Bakouch, Gabriel~Martin Blazquez,
  Guilherme Penedo, Lewis Tunstall, Andr{\'e}s Marafioti, Agust{\'\i}n~Piqueres
  Lajar{\'\i}n, Hynek Kydl{\'\i}{\v{c}}ek, Vaibhav Srivastav, Joshua Lochner,
  Caleb Fahlgren, Xuan~Son NGUYEN, Ben Burtenshaw, Cl{\'e}mentine Fourrier,
  Haojun Zhao, Hugo Larcher, Mathieu Morlon, Cyril Zakka, Colin Raffel,
  Leandro~Von Werra, and Thomas Wolf.
\newblock {Smol{LM}2: When Smol Goes Big {\textemdash} Data-Centric Training of
  a Fully Open Small Language Model}.
\newblock In \emph{Conference on Language Modeling}, 2025.

\bibitem[Loshchilov and Hutter(2019)]{loshchilov2017decoupled}
Ilya Loshchilov and Frank Hutter.
\newblock Decoupled weight decay regularization.
\newblock In \emph{International Conference on Learning Representations
  (ICLR)}, 2019.

\bibitem[Clark et~al.(2018)Clark, Cowhey, Etzioni, Khot, Sabharwal, Schoenick,
  and Tafjord]{clark2018arc}
Peter Clark, Isaac Cowhey, Oren Etzioni, Tushar Khot, Ashish Sabharwal, Carissa
  Schoenick, and Oyvind Tafjord.
\newblock Think you have solved question answering? try {ARC}, the {AI2}
  reasoning challenge.
\newblock \emph{arXiv preprint arXiv:1803.05457}, 2018.

\bibitem[Zellers et~al.(2019)Zellers, Holtzman, Bisk, Farhadi, and
  Choi]{zellers2019hellaswag}
Rowan Zellers, Ari Holtzman, Yonatan Bisk, Ali Farhadi, and Yejin Choi.
\newblock {H}ella{S}wag: Can a machine really finish your sentence?
\newblock In \emph{Proceedings of the 57th Annual Meeting of the {A}ssociation
  for {C}omputational {L}inguistics}, July 2019.

\bibitem[Bisk et~al.(2020)Bisk, Zellers, Bras, Gao, and Choi]{bisk2019piqa}
Yonatan Bisk, Rowan Zellers, Ronan~Le Bras, Jianfeng Gao, and Yejin Choi.
\newblock {PIQA: Reasoning about Physical Commonsense in Natural Language}.
\newblock In \emph{{AAAI}}, 2020.

\bibitem[Hendrycks et~al.(2021)Hendrycks, Burns, Basart, Zou, Mazeika, Song,
  and Steinhardt]{hendrycks2020mmlu}
Dan Hendrycks, Collin Burns, Steven Basart, Andy Zou, Mantas Mazeika, Dawn
  Song, and Jacob Steinhardt.
\newblock Measuring massive multitask language understanding.
\newblock In \emph{International Conference on Learning Representations
  (ICLR)}, 2021.

\bibitem[Hoffmann et~al.(2022)Hoffmann, Borgeaud, Mensch, Buchatskaya, Cai,
  Rutherford, de~las Casas, Hendricks, Welbl, Clark, Hennigan, Noland,
  Millican, van~den Driessche, Damoc, Guy, Osindero, Simonyan, Elsen, Vinyals,
  Rae, and Sifre]{hoffmann2022compute}
Jordan Hoffmann, Sebastian Borgeaud, Arthur Mensch, Elena Buchatskaya, Trevor
  Cai, Eliza Rutherford, Diego de~las Casas, Lisa~Anne Hendricks, Johannes
  Welbl, Aidan Clark, Tom Hennigan, Eric Noland, Katherine Millican, George
  van~den Driessche, Bogdan Damoc, Aurelia Guy, Simon Osindero, Karen Simonyan,
  Erich Elsen, Oriol Vinyals, Jack~William Rae, and Laurent Sifre.
\newblock An empirical analysis of compute-optimal large language model
  training.
\newblock In \emph{Conference on Neural Information Processing Systems
  (NeurIPS)}, 2022.

\bibitem[Sani et~al.(2025)Sani, Iacob, Cao, Lee, Marino, Gao, Zhao, Cai, Li,
  Qiu, and Lane]{sani2025photon}
Lorenzo Sani, Alex Iacob, Zeyu Cao, Royson Lee, Bill Marino, Yan Gao, Wanru
  Zhao, Dongqi Cai, Zexi Li, Xinchi Qiu, and Nicholas~D. Lane.
\newblock Photon: Federated {LLM} pre-training.
\newblock In \emph{Eighth Conference on Machine Learning and Systems}, 2025.

\bibitem[Charles and Konečný(2020)]{charles2020outsized}
Zachary Charles and Jakub Konečný.
\newblock On the outsized importance of learning rates in local update methods.
\newblock \emph{arXiv preprint arXiv:2007.00878}, 2020.

\bibitem[Belilovsky et~al.(2026)Belilovsky, Horoi, Oyallon, and
  Thérien]{belilovsky2026replica}
Eugene Belilovsky, Stefan Horoi, Edouard Oyallon, and Benjamin Thérien.
\newblock How inner optimizers shape replica scaling.
\newblock \emph{arXiv preprint arXiv:2610.19064}, 2026.

\bibitem[Su et~al.(2024)Su, Ahmed, Lu, Pan, Bo, and Liu]{su2021roformer}
Jianlin Su, Murtadha Ahmed, Yu~Lu, Shengfeng Pan, Wen Bo, and Yunfeng Liu.
\newblock {RoFormer}: Enhanced transformer with rotary position embedding.
\newblock \emph{Neurocomput.}, 2024.

\bibitem[Kumar et~al.(2025)Kumar, Owen, Chowdhury, and Güra]{kumar2025zclip}
Abhay Kumar, Louis Owen, Nilabhra~Roy Chowdhury, and Fabian Güra.
\newblock {ZClip}: Adaptive spike mitigation for {LLM} pre-training.
\newblock \emph{arXiv preprint arXiv:2504.02507}, 2025.

\bibitem[Semenov et~al.(2025)Semenov, Pagliardini, and
  Jaggi]{semenov2025benchmarking}
Andrei Semenov, Matteo Pagliardini, and Martin Jaggi.
\newblock Benchmarking {O}ptimizers for {L}arge {L}anguage {M}odel
  {P}retraining.
\newblock \emph{arXiv preprint arXiv:2509.01440}, 2025.

\bibitem[Sutskever et~al.(2013)Sutskever, Martens, Dahl, and
  Hinton]{sutskever2013momentum}
Ilya Sutskever, James Martens, George Dahl, and Geoffrey Hinton.
\newblock On the importance of initialization and momentum in deep learning.
\newblock In \emph{International Conference on Machine Learning (ICML)}, 2013.

\bibitem[H{\"a}gele et~al.(2024)H{\"a}gele, Bakouch, Kosson, Allal, von Werra,
  and Jaggi]{hagele2024scaling}
Alexander H{\"a}gele, Elie Bakouch, Atli Kosson, Loubna~Ben Allal, Leandro von
  Werra, and Martin Jaggi.
\newblock Scaling laws and compute-optimal training beyond fixed training
  durations.
\newblock In \emph{Conference on Neural Information Processing Systems
  (NeurIPS)}, 2024.

\bibitem[Douillard et~al.(2025)Douillard, Donchev, Rush, Kale, Charles, Teston,
  Garrett, Shen, McIlroy, Lacey, Rame, Szlam, Ranzato, and
  Barham]{douillard2025streaming}
Arthur Douillard, Yani Donchev, J~Keith Rush, Satyen Kale, Zachary Charles,
  Gabriel Teston, Zachary Garrett, Jiajun Shen, Ross McIlroy, David Lacey,
  Alexandre Rame, Arthur Szlam, MarcAurelio Ranzato, and Paul~R Barham.
\newblock Streaming {DiLoCo} with overlapping communication.
\newblock In \emph{Conference on Language Modeling}, 2025.

\bibitem[Sarfi et~al.(2026)Sarfi, Thérien, Lidin, and
  Belilovsky]{sarfi2025sparseloco}
Amir Sarfi, Benjamin Thérien, Joel Lidin, and Eugene Belilovsky.
\newblock Overcoming the communication-performance tradeoff in {LLM}
  pretraining.
\newblock \emph{arXiv preprint arXiv:2508.15706}, 2026.

\bibitem[Liu et~al.(2024)Liu, Chhaparia, Douillard, Kale, Rusu, Shen, Szlam,
  and Ranzato]{liu2024asynchronous}
Bo~Liu, Rachita Chhaparia, Arthur Douillard, Satyen Kale, Andrei~A. Rusu,
  Jiajun Shen, Arthur Szlam, and Marc'Aurelio Ranzato.
\newblock Asynchronous local-{SGD} training for language modeling.
\newblock \emph{arXiv preprint arXiv:2401.09135}, 2024.

\bibitem[Douillard et~al.(2026)Douillard, Rush, Donchev, Charles, Fallen,
  Dubey, Gog, Dean, Woodworth, Garrett, Keating, Bishop, Prior, Yvinec, Szlam,
  Ranzato, and Dean]{douillard2026decoupled}
Arthur Douillard, Keith Rush, Yani Donchev, Zachary Charles, Nova Fallen, Ayush
  Dubey, Ionel Gog, Josef Dean, Blake Woodworth, Zachary Garrett, Nate Keating,
  Jenny Bishop, Henry Prior, Edouard Yvinec, Arthur Szlam, Marc'Aurelio
  Ranzato, and Jeff Dean.
\newblock Decoupled {DiLoCo} for resilient distributed pre-training.
\newblock \emph{arXiv preprint arXiv:2604.21428}, 2026.

\bibitem[Ajanthan et~al.(2026)Ajanthan, Ramasinghe, Avraham, Dolatabadi,
  Koneputugodage, Shevchenko, Zuo, and Long]{ajanthan2026asyncmesh}
Thalaiyasingam Ajanthan, Sameera Ramasinghe, Gil Avraham, Hadi~Mohaghegh
  Dolatabadi, Chamin P~Hewa Koneputugodage, Violetta Shevchenko, Yan Zuo, and
  Alexander Long.
\newblock {AsyncMesh}: Fully asynchronous optimization for data and pipeline
  parallelism.
\newblock \emph{arXiv preprint arXiv:2601.22442}, 2026.

\bibitem[Koneputugodage et~al.(2026)Koneputugodage, Ajanthan, Ramasinghe,
  Dolatabadi, Siriwardhana, Avraham, Shevchenko, Pajak, Snewin, and
  Long]{koneputugodage2026factored}
Chamin P~Hewa Koneputugodage, Thalaiyasingam Ajanthan, Sameera Ramasinghe,
  Hadi~Mohaghegh Dolatabadi, Shamane Siriwardhana, Gil Avraham, Violetta
  Shevchenko, Karol Pajak, James Snewin, and Alexander Long.
\newblock Factored gossip {DiLoCo}: Reducing blocking communication in
  {DiLoCo}.
\newblock In \emph{International Conference on Machine Learning (ICML)}, 2026.

\bibitem[Sani et~al.(2026)Sani, Cao, Kurmanji, Iacob, Jovanovic, Gao, Zhao, and
  Lane]{sani2026fomoe}
Lorenzo Sani, Zeyu Cao, Meghdad Kurmanji, Alex Iacob, Andrej Jovanovic, Yan
  Gao, Wanru Zhao, and Nicholas~D. Lane.
\newblock {FoMoE}: Breaking the full-replica barrier with a federation of
  {MoEs}.
\newblock \emph{arXiv preprint arXiv:2606.19025}, 2026.

\bibitem[McCandlish et~al.(2018)McCandlish, Kaplan, Amodei, and {OpenAI Dota
  Team}]{mccandlish2018batch}
Sam McCandlish, Jared Kaplan, Dario Amodei, and {OpenAI Dota Team}.
\newblock An empirical model of large-batch training.
\newblock \emph{arXiv preprint arXiv:1812.06162}, 2018.

\bibitem[Zhang et~al.(2025)Zhang, Morwani, Vyas, Wu, Zou, Ghai, Foster, and
  Kakade]{zhang2025critical}
Hanlin Zhang, Depen Morwani, Nikhil Vyas, Jingfeng Wu, Difan Zou, Udaya Ghai,
  Dean Foster, and Sham~M. Kakade.
\newblock How does critical batch size scale in pre-training?
\newblock In \emph{International Conference on Learning Representations
  (ICLR)}, 2025.

\bibitem[Liu et~al.(2025)Liu, Su, Yao, Jiang, Lai, Du, Qin, Xu, Lu, Yan, Chen,
  Zheng, Liu, Liu, Yin, He, Zhu, Wang, Wang, Dong, Zhang, Kang, Zhang, Xu,
  Zhang, Wu, Zhou, and Yang]{liu2025muon}
Jingyuan Liu, Jianlin Su, Xingcheng Yao, Zhejun Jiang, Guokun Lai, Yulun Du,
  Yidao Qin, Weixin Xu, Enzhe Lu, Junjie Yan, Yanru Chen, Huabin Zheng, Yibo
  Liu, Shaowei Liu, Bohong Yin, Weiran He, Han Zhu, Yuzhi Wang, Jianzhou Wang,
  Mengnan Dong, Zheng Zhang, Yongsheng Kang, Hao Zhang, Xinran Xu, Yutao Zhang,
  Yuxin Wu, Xinyu Zhou, and Zhilin Yang.
\newblock Muon is scalable for {LLM} training.
\newblock \emph{arXiv preprint arXiv:2502.16982}, 2025.

\bibitem[Qiu et~al.(2025)Qiu, Chen, Phan, Lei, and
  Wilson]{qiu2025matrixpreconditioned}
Shikai Qiu, Zixi Chen, Hoang Phan, Qi~Lei, and Andrew~Gordon Wilson.
\newblock Hyperparameter transfer enables consistent gains of
  matrix-preconditioned optimizers across scales.
\newblock In \emph{Conference on Neural Information Processing Systems
  (NeurIPS)}, 2025.

\bibitem[Wen et~al.(2026)Wen, Hall, Ma, and Liang]{wen2026fantastic}
Kaiyue Wen, David Leo~Wright Hall, Tengyu Ma, and Percy Liang.
\newblock Fantastic pretraining optimizers and where to find them.
\newblock In \emph{International Conference on Learning Representations
  (ICLR)}, 2026.

\end{thebibliography}
\endgroup

\appendix
\etocdepthtag{appendix}
\newcommand{\appendixcontents}{%
  \begingroup
    \normalsize
    \setlength{\baselineskip}{13pt}
    \setlength{\parskip}{0pt}
    \etocsettagdepth{main}{none}
    \etocsettagdepth{appendix}{subsubsection}
    \etocsetstyle{section}{}{}{%
      \par\addvspace{7pt}\noindent
      {\bfseries\etocnumber\quad\etocname\hfill\etocpage\par}%
      \nopagebreak[4]}{}
    \etocsetstyle{subsection}{}{}{%
      \noindent\hspace*{1.5em}\etocnumber\quad\etocname
      \nobreak\leaders\hbox to .8em{\hss.\hss}\hfill\nobreak\etocpage\par}{}
    \etocsettocstyle{\section*{Appendix}}{}
    \tableofcontents
  \endgroup
}
\makeatletter
\ifpaper@twocolumn
  \twocolumn[\begin{@twocolumnfalse}\appendixcontents\vspace{8pt}\end{@twocolumnfalse}]
\else
  \appendixcontents
\fi
\makeatother

\raggedbottom
\pretocmd{\section}{\FloatBarrier}{}{}
\clubpenalty=10000
\widowpenalty=10000
\setcounter{topnumber}{4}
\setcounter{bottomnumber}{3}
\setcounter{totalnumber}{6}
\setcounter{dbltopnumber}{4}
\renewcommand{\topfraction}{.95}
\renewcommand{\bottomfraction}{.95}
\renewcommand{\textfraction}{.05}
\renewcommand{\floatpagefraction}{.9}
\renewcommand{\dbltopfraction}{.95}
\renewcommand{\dblfloatpagefraction}{.75}
\setlength{\floatsep}{6pt plus 2pt minus 1pt}
\setlength{\textfloatsep}{6pt plus 2pt minus 1pt}
\setlength{\dblfloatsep}{6pt plus 2pt minus 1pt}
\setlength{\dbltextfloatsep}{6pt plus 2pt minus 1pt}
\makeatletter
\setlength{\@fptop}{0pt}
\setlength{\@fpsep}{7pt plus 2pt minus 2pt}
\setlength{\@fpbot}{0pt plus 1fil}
\setlength{\@dblfptop}{0pt}
\setlength{\@dblfpsep}{7pt plus 2pt minus 2pt}
\setlength{\@dblfpbot}{0pt plus 1fil}
\makeatother

\setlength{\intextsep}{6pt plus 2pt minus 1pt}

\setlength{\parskip}{2pt plus 1pt minus 1pt}
\setlength{\abovedisplayskip}{5pt plus 2pt minus 1pt}
\setlength{\belowdisplayskip}{5pt plus 2pt minus 1pt}
\setlength{\abovedisplayshortskip}{3pt plus 1pt minus 1pt}
\setlength{\belowdisplayshortskip}{4pt plus 1pt minus 1pt}

\makeatletter
\newenvironment{appendixfloattable}[1][htbp]
  {\ifpaper@twocolumn \@nameuse{table*}[#1]\else \@nameuse{table}[#1]\fi}
  {\ifpaper@twocolumn \@nameuse{endtable*}\else \@nameuse{endtable}\fi}
\makeatother
\renewenvironment{widetable}[1][htbp]
  {\par\addvspace{2pt}\noindent\begin{minipage}{\linewidth}\captionsetup{type=table,skip=4pt}}
  {\end{minipage}\par\addvspace{2pt}}

\captionsetup[figure]{skip=4pt}

\Needspace{12\baselineskip}
\section{Experimental Setup}
\label{app:experimental-setup}
\subsection{Model configurations and parameterization}
\label{app:model-configurations}
\label{app:model-size-protocol}
Our experiments on language modelling use the pre-norm architectures specified in
\cref{tab:model-configurations}. All our width and depth multiples are calculated relative to
\smallmodel. All models use multi-head attention, full RoPE~\citep{su2021roformer}, gated SiLU
feed-forward layers, affine bias-free LayerNorm, and untied embeddings.
Embedding, pre-block and per-head Q/K normalization are enabled.
We fix head dimension across widths, use equal query and key/value head
counts, and apply inverse-square-root attention scaling.
\paragraph{Tensor factors.}
For width and depth scaling, each worker applies the \completep initialization, forward and optimizer
factors \citep{dey2025completep}, composed with the local-batch
and token factors of \completedp~\citep{mlodozeniec2025completedp}
as described in \cref{sec:transfer-recipe}.
\begin{widetable}[!htbp]
 \centering\small
 \setlength{\tabcolsep}{8pt}
 \renewcommand{\arraystretch}{1.0}
 \caption{\textbf{Model architectures and reference configurations.}
 Gradient norm specifies the clipping limit used with \zclip~\citep{kumar2025zclip}.
 Batch counts sequences. Steps include warmup.
 Tokens are continuation budgets in billions. Width/depth sweeps retain
 the \smallmodel continuation budget.}
 \label{tab:model-configurations}
 \label{tab:training-budgets}
 \begin{tabular}{@{}p{.43\linewidth}*{3}{>{\raggedleft\arraybackslash}p{\dimexpr.19\linewidth-2\tabcolsep\relax}}@{}}
 \toprule
 \textbf{Setting} & \textbf{\smallmodel} & \textbf{\mediummodel} & \textbf{\largemodel}\\
 \midrule
 \multicolumn{4}{@{}l@{}}{\textbf{Architecture}}\\
 Parameters (M) & \(29.2\) & \(162.3\) & \(822.9\)\\
 Blocks & \(4\) & \(12\) & \(12\)\\
 Width & \(256\) & \(768\) & \(2{,}048\)\\
 Attention heads & \(4\) & \(12\) & \(32\)\\
 \addlinespace[5pt]
 \multicolumn{4}{@{}l@{}}{\textbf{Training}}\\
 Gradient norm & \(1.0\) & \(0.5\) & \(0.1\)\\
 Batch (sequences) & \(64\) & \(256\) & \(512\)\\
 Steps & \(6{,}144\) & \(6{,}144\) & \(16{,}384\)\\
 Continuation tokens (B) & \(0.537\) & \(2.147\) & \(15.032\)\\
 \end{tabular}\par\nointerlineskip\vspace{5pt}
 \begin{tabular*}{\linewidth}{@{\extracolsep{\fill}}lrlr@{}}
 \midrule
 \multicolumn{4}{@{}l@{}}{\textbf{Shared across all models}}\\
 Vocabulary size & \(50{,}368\) & Sequence length & \(2{,}048\)\\
 Head dimension & \(64\) & FF expansion (nominal) & \(4\)\\
 RoPE base & \(10{,}000\) & Initialization std. dev. & \(0.02\)\\
 Warmup steps & \(2{,}048\) & &\\
 \bottomrule
 \end{tabular*}
\end{widetable}

\subsection{Training hyperparameters and evaluation}
\label{app:common-protocol}
\label{app:experiment-protocols}
We hold data and optimization settings fixed within each comparison
(\cref{tab:training-hyperparameters,tab:shared-settings,tab:training-budgets}). For sequence
length \(S\) and global sequence batch \(B_{\mathrm{seq}}\),
\(B=SB_{\mathrm{seq}}\), \(b=B/M\), and a continuation budget \(D\)
gives \(D/B\) local updates per worker.
Validation loss is token-weighted cross-entropy in nats, excluding training
z-loss. Final evaluations follow the last synchronization.
\begin{appendixfloattable}[!htbp]
 \centering\small
 \setlength{\tabcolsep}{5pt}
 \caption{\textbf{Warmup and continuation learning rates.}
 Rates precede tensor-specific multipliers. Averaging and \nesterov share
 the warmup in each row.}
 \label{tab:training-hyperparameters}
 \label{tab:training-schedules}
 \begin{tabular}{@{}lrrrr@{}}
 \toprule
 & Warmup & Averaging & \multicolumn{2}{c}{\nesterov}\\
 \cmidrule(l){4-5}
 Model / recipe & Inner rate & Inner rate & Inner rate & Outer rate\\
 \midrule
 \smallmodel & \(.016\) & \(.008\) & \(.016\) & \(.1\)\\
 \mediummodel & \(.032\) & \(.016\) & \(.016\) & \(.2\)\\
 \midrule
 \multicolumn{5}{@{}l@{}}{\textbf{\largemodel}}\\
 \dilocofit & \(.000417\) & \(.000417\) & \(.000417\) & \(.6\)\\
 \method{} (Reference) & \(.00581\) & \(.0164\) & \(.0164\) & \(.6\)\\
 \method{} (Small-tuned) & \(.016\) & \(.0226\) & \(.0453\) & \(.1\)\\
 \bottomrule
 \end{tabular}
\end{appendixfloattable}

Reference batch sizes, gradient-norm limits and warmup length follow
\lordo~\citep{jovanovic2026lordo}, drawing on the optimizer
benchmarks of \citet{semenov2025benchmarking}.
The \adam moment coefficients and base epsilon use the defaults of
\citet{kingma2015adam}. We initialise the \nesterov~\citep{sutskever2013momentum} momentum coefficient with that of
\citet{charles2025dilocoscaling}. We also adopt their initial outer-rate grid.

\paragraph{Warmup and continuation.}
For RQ1--RQ3, we increase the learning rate linearly during \ddp warmup
and keep it constant during continuation. For width and depth sweeps,
we run warmup at each candidate rate and share the resulting checkpoints
across outer optimizers. For batch-size and local-step sweeps, we start
from one reference checkpoint per model and change the rate after warmup,
following
\mtdao~\citep{iacob2025mtdao}.
Each continuation restores the inner moments, their step counts and clipping
history. Outer-optimizer states start at zero. We keep sample order fixed
across worker counts.
\phantomsection
\label{app:rq4-large-protocol}
For the \largemodel model runs in RQ4, a constant-rate continuation ends with the cooldown in
\cref{tab:shared-settings}~\citep{hagele2024scaling}.
Both recipes use local-batch factor \(\sqrt8\) and no token-budget correction.

\begin{widetable}[!htbp]
 \renewcommand{\arraystretch}{1.0}
 \begin{minipage}[t]{.66\linewidth}
 \centering\small\setlength{\tabcolsep}{3pt}
 \caption{\textbf{Shared training and optimizer settings.}}
 \label{tab:shared-settings}
 \begin{tabular}{@{}ll@{}}
 \toprule
 Setting & Value\\
 \midrule
 Residual-depth exponent & \(\alpha=1\)\\
 Workers / local steps & \(M=4\), \(K=32\)\\
 Precision / seed & BF16 / \(42\)\\
 Inner \adam betas / base epsilon & \((0.9,0.999)\) / \(10^{-8}\)\\
 Weight decay / z-loss coefficient & \(0\) / \(10^{-5}\)\\
 Averaging rate / momentum & \(1\) / \(0\)\\
 \nesterov momentum & \(0.9\)\\
 Outer \adam rate / betas / base \(\epsilon\) & \(0.02\) / \((0.9,0.999)\) / \(10^{-8}\)\\
 Large cooldown & \(2{,}048\) steps\\
 \bottomrule
 \end{tabular}
 \end{minipage}\hfill
 \begin{minipage}[t]{.31\linewidth}
 \centering\small\setlength{\tabcolsep}{3pt}
 \caption{\textbf{Data and validation.}}
 \label{tab:data-validation}
 \begin{tabular}{@{}lr@{}}
 \toprule
 Source & Share\\
 \midrule
 FineWeb-Edu & \(70\%\)\\
 Cosmopedia & \(10\%\)\\
 Python-Edu & \(10\%\)\\
 FineMath-4plus & \(5\%\)\\
 InfiWebMath-4plus & \(5\%\)\\
 \midrule
 Validation & Tokens\\
 \midrule
 Checkpoints & \(21.0\)M\\
 Final & \(104.8\)M\\
 \bottomrule
 \end{tabular}
 \end{minipage}
\end{widetable}

\paragraph{Sweep axes.}
\Cref{tab:sweep-grids} lists the learning rates and configurations tested.
For the \smallmodel \nesterov reference, we test every combination of
inner and outer rates. We extend the outer-rate grid from
\citet{charles2025dilocoscaling} with \(0.05\) and \(0.1\)
to bracket the minimum. Batch sweeps cover both unchanged and square-root
learning-rate scaling.

\begin{widetable}[!htbp]
 \centering\small
 \renewcommand{\arraystretch}{1.0}
 \caption{\textbf{Sweep coordinates for transfer measurements.}
 Each batch panel specifies which coordinate varies. The listed values
 summarize the separate sweeps. For joint tuning across local steps,
 \(\gamma_0=0.1\) for \smallmodel and \(0.2\) for \mediummodel.}
 \label{tab:sweep-grids}
 \begin{tabular}{@{}ll@{}}
 \toprule
 Sweep & Grid\\
 \midrule
 Width and depth multiples & \(1,2,4,8,16\)\\
 Reference and width/depth inner rate & \(0.002,0.004,0.008,0.016,0.032,0.064\)\\
 Reference \nesterov outer rate & \(0.05,0.1,0.2,0.4,0.6,0.8,1.0\)\\
 Conditional outer learning rate & \(0.05,0.1,0.2,0.4\)\\
 Worker count & \(2,4,8\)\\
 Local batch multiples, averaging & \(1,2,4,8\) (\smallmodel), \(1,2,4\) (\mediummodel)\\
 Local batch multiples, \nesterov & \(1,2,4\)\\
 Local steps per round & \(K=1,2,4,8,16,32,64\)\\
 Reference continuation steps & \(4{,}096\)\\
 Joint inner learning rate & \(0.008,0.016,0.032\)\\
 Joint outer learning rate & \(\gamma_0\times\{1/4,1/2,1,2,4\}\)\\
 \bottomrule
 \end{tabular}
\end{widetable}

\section{Additional Experimental Results}
\label{app:additional-results}

\subsection{Outer-rate transfer across model sizes}
\label{app:outer-protocol}
We select an inner learning rate for each model size and parameterization
at the reference outer rate, then hold it fixed while sweeping the outer
rate (\cref{fig:outer-rate-transfer}).

\begin{widefigure}[!htbp]
 \centering
 \begin{subfigure}[t]{.46\linewidth}\centering
  \captionsetup{position=top,skip=3pt}\caption{Width}\label{fig:rq1-outer-width}
  \includegraphics[width=\linewidth]{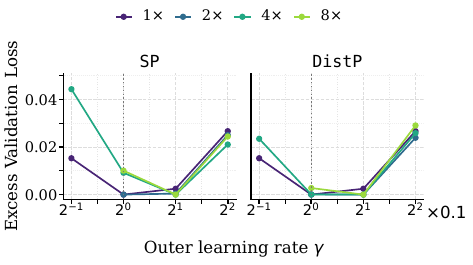}
 \end{subfigure}\hfill
 \begin{subfigure}[t]{.46\linewidth}\centering
  \captionsetup{position=top,skip=3pt}\caption{Depth}\label{fig:rq1-outer-depth}
  \includegraphics[width=\linewidth]{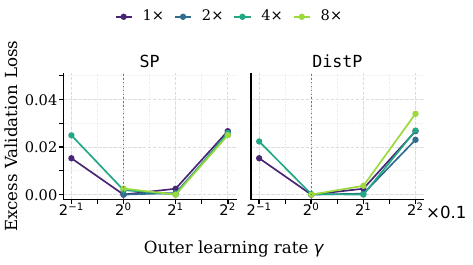}
 \end{subfigure}
 \caption{\textbf{\nesterov's reference outer rate remains effective across model sizes.}
 Width and depth through \(8\times\), relative to \(256\) and four layers,
 at \(M=4,K=32\). Inner rates are selected separately
 at each model size at outer rate \(.1\). Loss is relative to each curve's best tested outer rate.}
 \label{fig:outer-rate-transfer}
\end{widefigure}

\FloatBarrier
\Needspace*{12\baselineskip}
\subsection{Depth transfer curves}
\label{app:depth-curves}
\Cref{fig:rq1-depth} gives the depth results for \cref{sec:rq1,sec:rq3}.
We use the width-sweep protocol.

\begin{widefigure}[!htbp]
 \centering
 \begin{minipage}{.86\linewidth}
 \rqmatrixpanels{depth}{fig:rq1-depth}{1}
 \end{minipage}
 \caption{\textbf{Depth transfer extends to all three outer optimizers.}
 Rows show averaging, \nesterov and \adam at \(M=4,K=32\).
 Colors denote depth through \(16\times\) relative to four layers at width \(256\).
 Points show validation loss across inner rates.
 The reference rate \(0.016\) remains near-optimal.}
 \label{fig:rq1-depth}
\end{widefigure}

\subsection{Depth ablation without Q/K normalization}
\label{app:model-size-curves}
\Cref{fig:rq1-qk-off-depth} shows the Q/K normalization ablation
discussed in \cref{sec:rq1}.

\begin{widefigure}[!htbp]
 \centering
 \includegraphics[width=.84\linewidth]{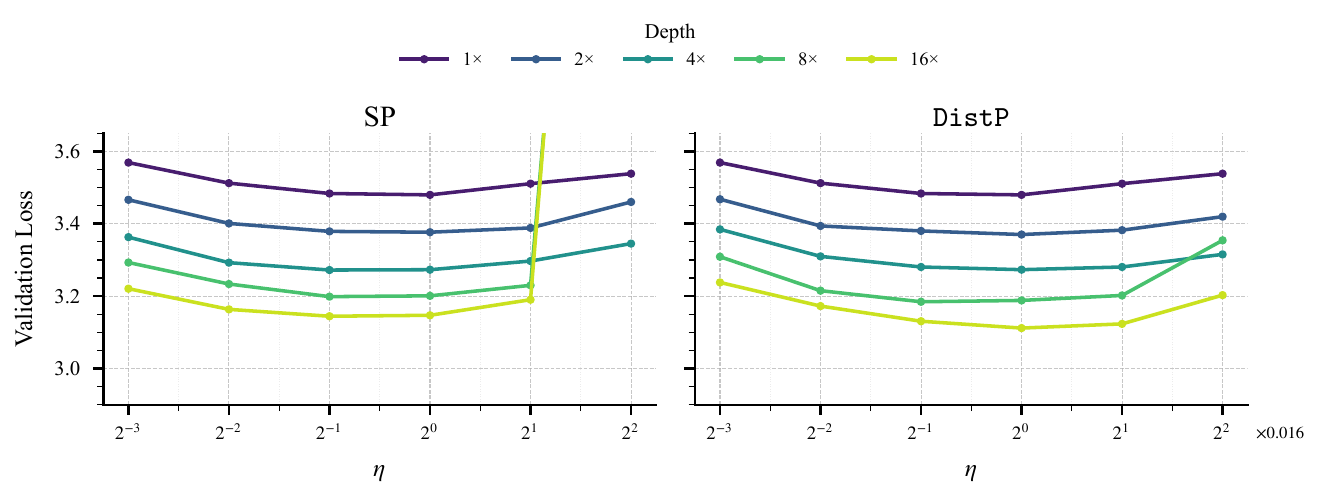}
 \caption{\textbf{Approximate depth transfer persists without Q/K normalization.}
 Averaging at \(M=4,K=32\), with embedding normalization and pre-norm
 blocks retained. Colors show depth through \(16\times\) relative to four layers at width \(256\).}
 \label{fig:rq1-qk-off-depth}
\end{widefigure}

\FloatBarrier
\Needspace{6\baselineskip}
\subsection{Transfer across model and training configurations}
\label{app:small-transfer-regimes}
\Cref{fig:rq2-small-regimes,fig:rq3-small-regimes} show the \smallmodel
results corresponding to RQ2 and RQ3. With \nesterov, \smallmodel favors
an unchanged inner learning rate when adding workers and less than
square-root scaling when increasing local batch size. In comparison,
\mediummodel favors higher rates. Each comparison uses the model's
reference outer learning rate.
\begin{widefigure}[!htbp]
 \centering
 \renewcommand{\rqcadenceheading}{Inner learning rate transfer}
 \begin{minipage}{.9\linewidth}
 \rqregimepanels{rq2-averaging-small-regimes}{fig:rq2-small-workers}{fig:rq2-small-local-batch}{fig:rq2-small-local-steps}
 \end{minipage}
 \caption{\textbf{Small-model averaging supports the same transfer rules.}
 \smallmodel (\(29.2\)M) at fixed tokens, with reference \(M_0=4\),
 \(b_0=16\) sequences, \(K=32\), \(\eta_0=.008\).
 (a,b) Guides mark \(\eta_0\) and \(\eta_0\sqrt{b/b_0}\). The dotted path
 joins best tested rates in (b).
 (c) Half, reference and double inner rates at fixed global batch.
 Excess loss is relative to each \(K\)'s best tested rate.}
 \label{fig:rq2-small-regimes}
\end{widefigure}
\begin{widefigure}[!htbp]
 \centering
 \begin{minipage}{.9\linewidth}
 \rqregimepanels{rq3-nesterov-small-regimes}{fig:rq3-small-workers}{fig:rq3-small-local-batch}{fig:rq3-small-local-steps}
 \end{minipage}
 \caption{\textbf{\nesterov's batch preferences depend on the model and recipe.}
 \smallmodel (\(29.2\)M) at fixed tokens, with reference \(M_0=4\),
 \(b_0=16\) sequences, \(K=32\), \(\eta_0=.016\), \(\gamma_0=.1\).
 (a,b) Inner-rate sweeps at \(\gamma_0\), with guides at \(\eta_0\) and
 \(\eta_0\sqrt{b/b_0}\). The dotted path joins best tested rates in (b).
 (c) Half, reference and double outer rates at \(\eta_0\) and fixed global batch.
 Excess loss is relative to each \(K\)'s best tested outer rate.}
 \label{fig:rq3-small-regimes}
\end{widefigure}

\subsection{Joint inner--outer learning-rate tuning}
\label{app:joint-retuning}
The main comparisons vary one learning rate at a time.
\Cref{fig:joint-retuning} shows how the outer rate changes the effect of
local-batch scaling, and how tuning both rates can improve on outer-only
tuning across synchronization intervals.

\begin{widefigure}[!htbp]
 \centering
 \includegraphics[width=.9\linewidth]{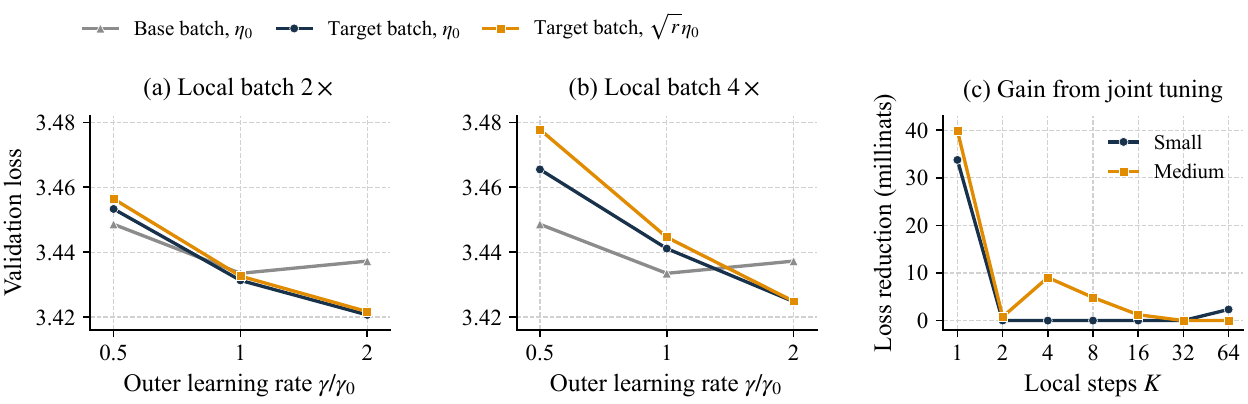}
 \caption{\textbf{Joint tuning adds gains beyond the fixed-rate comparisons.}
 (a,b) \smallmodel \nesterov at \(M=4,K=32\), with
 \(\eta_0=.016\), \(\gamma_0=.1\) and local-batch multiple \(r=b/b_0\).
 Larger outer rates reduce the loss penalty from square-root inner scaling.
 (c) Validation-loss reduction from tuning both rates relative to tuning
 only the outer rate at \(\eta=.016\), using each interval's best tested
 settings. Global batch and tokens are fixed within each model's cadence sweep.}
 \label{fig:joint-retuning}
\end{widefigure}

\section{Model Quality Across Size and Compute}
\label{app:conditional-scaling}

\label{app:compute-frontier}
\paragraph{Measurements and compute.}
The comparisons in \cref{sec:rq4} select learning rates from the six-rate
grid in \cref{tab:sweep-grids}. For each of the eight non-reference
architectures, we evaluate seven checkpoints at steps
\(2{,}560+512j\), \(j=0,\ldots,6\). The reference model and separate
\largemodel runs are excluded.
We estimate cumulative training FLOPs as
\begin{equation}
 C=3D\bigl[\ell(8w^2+6wh+2Sw)+2wV\bigr],
\end{equation}
where \(w\) is width, \(\ell\) depth, \(h\) rounded SwiGLU hidden width,
\(S\) sequence length, and \(V\) vocabulary size.
The attention term uses a causal approximation. This estimates forward
and backward matrix-multiplication FLOPs. Global tokens \(D\) include
warmup and sum over workers.

\paragraph{Loss--compute fits.}
We fit the best observed loss at or below each common FLOP budget using
the power law in \cref{tab:compute-fit}. We use Huber loss on log-loss
residuals with threshold \(10^{-3}\), over the same compute range for both
parameterizations. To assess how individual architectures affect the fit,
we omit each architecture in turn and refit the resulting envelope.

\begin{widetable}[!htbp]
 \centering\small
 \caption{\textbf{Compute-frontier fits summarize the persistent quality gain.}
 Coefficients of \(L(C)=E+A(C/10^{18})^{-\alpha}\), with \(C\) in training FLOPs.
 Exponent ranges come from refitting after omitting each architecture in turn.}
 \label{tab:compute-fit}
 \begin{tabular}{@{}llrrrr@{}}
 \toprule
 Outer & Method & \(E\) & \(A\) & \(\alpha\) & \(\alpha\) range\\
 \midrule
 Averaging & \standardp & \(2.937\) & \(0.135\) & \(0.493\) & \([0.47,0.57]\)\\
 & \method & \(2.708\) & \(0.304\) & \(0.327\) & \([0.31,0.38]\)\\
 \midrule
 \nesterov & \standardp & \(2.919\) & \(0.133\) & \(0.494\) & \([0.48,0.58]\)\\
  & \method & \(2.700\) & \(0.287\) & \(0.342\) & \([0.31,0.35]\)\\
 \bottomrule
 \end{tabular}
\end{widetable}

\begin{widefigure}[!htbp]
 \centering
 \begin{minipage}{.95\linewidth}
 \begin{subfigure}[t]{.5247\linewidth}\centering
  \captionsetup{position=top,skip=3pt}\caption{Width and depth scaling}
  \label{fig:rq4-nesterov-compute-trajectories}
  \includegraphics[width=\linewidth]{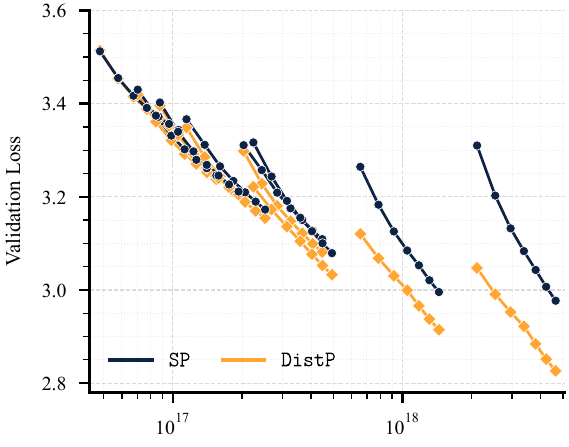}
 \end{subfigure}\hfill
 \begin{subfigure}[t]{.4553\linewidth}\centering
  \captionsetup{position=top,skip=3pt}\caption{Loss--compute fit}
  \label{fig:rq4-nesterov-compute-frontier}
  \includegraphics[width=\linewidth]{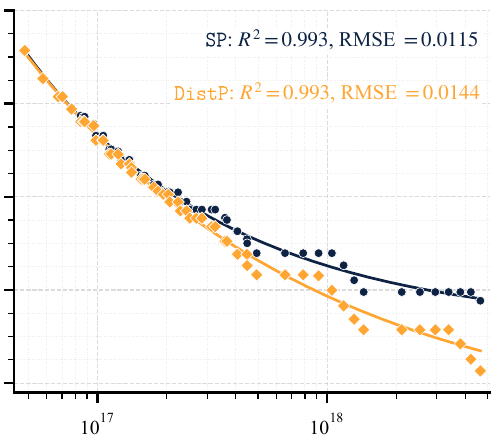}
 \end{subfigure}\par\vspace{1pt}FLOPs
 \end{minipage}
 \caption{\textbf{\method also improves loss at matched compute under \nesterov.}
 Outer learning rate is \(0.1\).
 (a) Validation loss using each architecture and parameterization's best-tested
 inner learning rate at each measured compute budget. (b) Best observed loss at or below each FLOP budget,
 fitted with \(L(C)=E+A(C/10^{18})^{-\alpha}\). Annotations give in-sample \(R^2\) and RMSE.}
 \label{fig:rq4-scaling-nesterov}
\end{widefigure}

\paragraph{Large-model evaluation.}
We evaluate downstream tasks with LightEval
(\cref{fig:rq4-downstream-trajectories}), using character-normalized
accuracy for ARC, HellaSwag and PIQA, and question-weighted accuracy for
MMLU. The reported mean weights the five tasks equally.

\begin{widefigure}[!htbp]
 \centering
 \includegraphics[width=.77\linewidth]{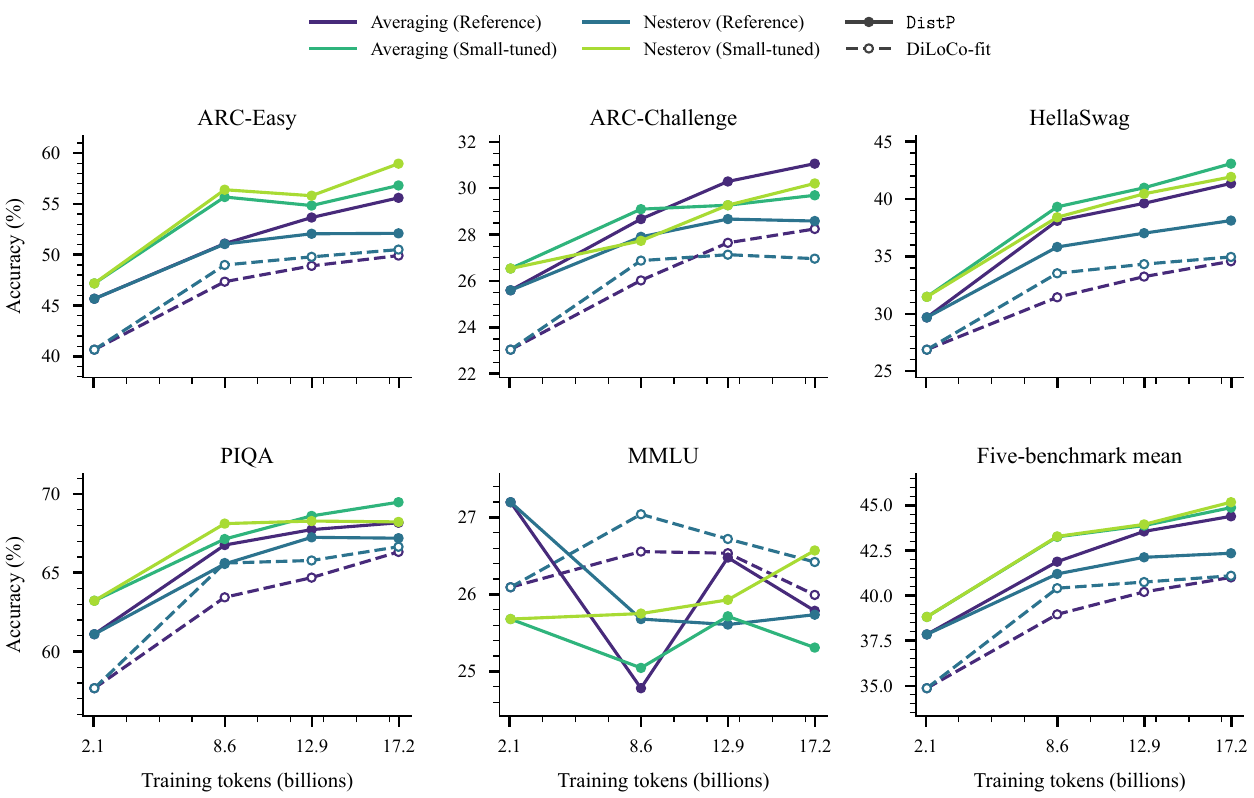}
 \caption{\textbf{Transferred recipes improve downstream accuracy.}
 \largemodel task scores, zero-shot except five-shot MMLU.
 The final panel averages the five tasks.}
 \label{fig:rq4-downstream-trajectories}
\end{widefigure}

\section{Feature-scaling checks under averaging}
\label{app:toy-averaging}
\begin{appendixfloattable}[!htbp]
 \centering\small
 \renewcommand{\arraystretch}{1.05}
 \caption{\textbf{Protocols for the feature and coordinate diagnostics.}
 Both use \(M\in\{1,4\}\), \(K=32\), constant learning rates and no warmup.}
 \label{tab:coordinate-protocol}
 \begin{tabular}{@{}lll@{}}
 \toprule
 Setting & MNIST & Shakespeare\\
 \midrule
 Width & \(64,128,256,512\) at \(L=4\) & \(256\)\\
 Depth & \(2,4,\ldots,256\) at \(N=256\) & \(2,4,8,16,32,64\)\\
 Residual exponents & \(0.5,1\) (\(1\) for width) & \(0.5,1\)\\
 Communication rounds & \(1\) & \(2\)\\
 Paired seeds & \(50\) & \(3\)\\
 Local batch & \(128\) examples & \(8\) sequences of \(1{,}024\) tokens\\
 Probe & First \(128\) training examples & \(2\) fixed training sequences\\
 Precision & FP64 after FP32 initialization & FP32\\
 Optimizer / betas & \adam / \((0.9,0.999)\) & \adamw / \((0.9,0.95)\)\\
 Base learning rate / epsilon & See formulas below & \(10^{-3}\) / \(10^{-12}\)\\
 Weight decay / clipping & \(0\) / none & \(0.1\) / \(1\)\\
 \bottomrule
 \end{tabular}
\end{appendixfloattable}

\subsection{MNIST branch diagnostics}

\paragraph{Setup.}
We use \completep's MNIST residual network and initialization
\citep{dey2025completep}, with branch
multiplier \(\frac{1}{\left(\frac{L}{2}\right)^\alpha}\). Following the released \completep
linearization notebook, we train only the first matrix of the second residual
branch and freeze all other parameters. Workers start from the same
parameters and receive consecutive batches in round-robin order.
\Cref{tab:coordinate-protocol} gives the settings for both diagnostics.
The local optimizer uses
\(\eta=10^{-4}N^{-1/2}(L/2)^{\alpha-1}\) and
\(\epsilon=10^{-8}(N/256)^{-2}(L/2)^{-\alpha}\).

\paragraph{Diagnostics.}
To separate nonlinear error from cancellation, let \(\theta_0\) be the
trained matrix at initialization and \(H(\theta)\)
the branch's probe output before residual scaling. Write
\(\delta\theta^m=\theta^m_{0,32}-\theta_0\),
\(\overline{\delta\theta}=\frac{1}{M}\sum_m\delta\theta^m\), and \(J=DH(\theta_0)\).
For \(\Delta H=H(\theta_0+\overline{\delta\theta})-H(\theta_0)\), define
\begin{equation}
 R=\frac{\|\Delta H-J\overline{\delta\theta}\|_{\mathrm{RMS}}}
             {\|J\overline{\delta\theta}\|_{\mathrm{RMS}}},\qquad
 C=\frac{\|J\overline{\delta\theta}\|_{\mathrm{RMS}}}
             {\frac{1}{M}\sum_m\|J\delta\theta^m\|_{\mathrm{RMS}}}.
 \label{eq:toy-remainder-cancellation}
\end{equation}
Here \(R\) measures relative linearization error and \(C\) the fraction
of the mean worker tangent magnitude retained after averaging.
We also report RMS feature magnitudes and changes.

\begin{widefigure}[!htbp]
 \centering
 \includegraphics[width=0.74\textwidth]{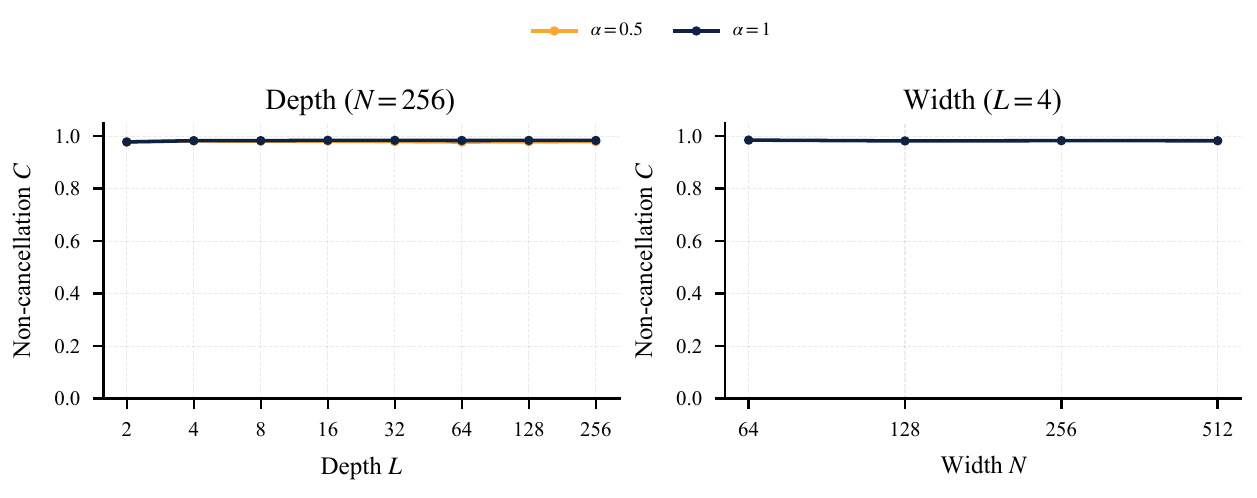}
 \caption{\textbf{Averaging retains about \(98\%\) of worker tangent magnitude.}
 Non-cancellation ratio \(C\) after \(32\) local steps with four workers,
 across depth and width. Medians and interquartile ranges over \(50\) seeds
 assess the condition in \cref{app:server-averaging-scaling} for the MNIST branch.}
 \label{fig:toy-averaging}
\end{widefigure}

\begin{widefigure}[!htbp]
 \centering
 \includegraphics[width=.90\textwidth]{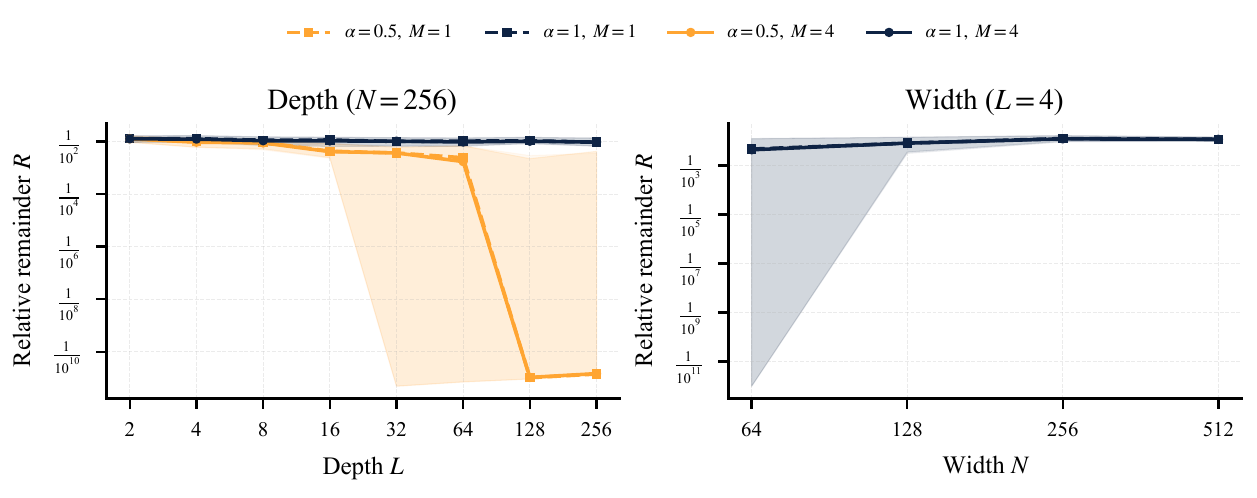}
 \caption{\textbf{Across depth, nonlinear remainders stay near \(1\%\) for \(\alpha=1\)
 and approach roundoff for \(\alpha=0.5\).}
 MNIST branch: relative distance to linearization \(R\) across depth and width after
 \(32\) local steps. Dashed squares: single-worker control (\(M=1\)).
 Solid circles: averaging control (\(M=4\)).
 Lines and bands show medians and interquartile ranges over \(50\) seeds.
 Values below \(\frac{1}{10^{12}}\) are clipped for log display.}
 \label{fig:toy-linearisation}
\end{widefigure}

\begin{widefigure}[!htbp]
 \centering
 \includegraphics[width=0.75\textwidth]{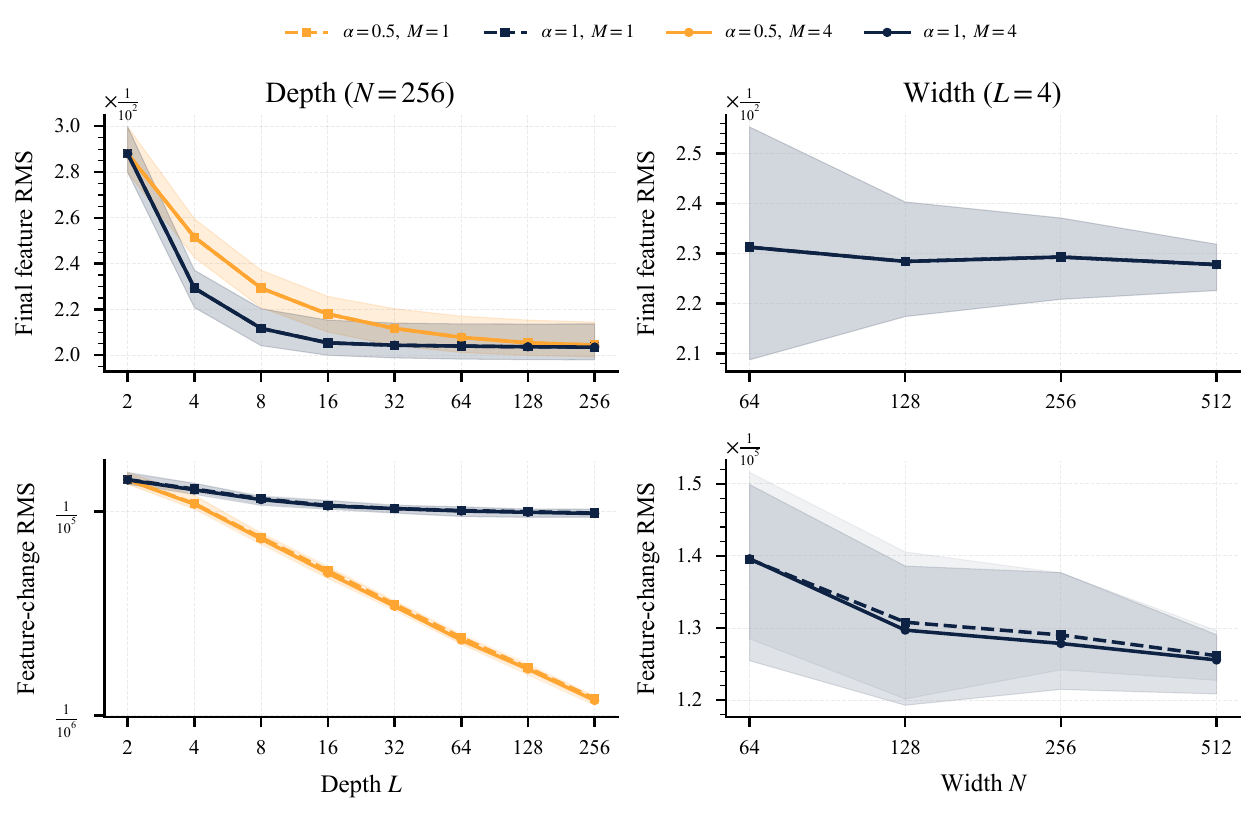}
 \caption{\textbf{Single-worker and averaging controls have similar feature scales.}
 MNIST branch RMS (top) and change RMS (bottom) before residual scaling,
 after \(32\) local steps. Dashed squares: \(M=1\). Solid circles: \(M=4\) with averaging.
 Depth varies at \(N=256\), and width at \(L=4\). Lines/bands: medians/interquartile ranges over \(50\) seeds.}
 \label{fig:toy-coordinates}
\end{widefigure}

\FloatBarrier
\subsection{Shakespeare transformer coordinate checks}
\label{app:shakespeare-coordinates}

We extend \completep's Shakespeare coordinate check~\citep{dey2025completep}
to two rounds of local training and averaging, using the settings in
\cref{tab:coordinate-protocol}.

\paragraph{Model and optimization.}
We use the character-level GPT from \texttt{EleutherAI/nanoGPT-mup},
retaining its architecture, learning-rate and \adamw epsilon rules.
The model uses learned positional embeddings and tied input/output
embeddings, without Q/K or embedding normalization. We train all parameters.
The depth-scaled variants use the same
residual multiplier as the MNIST diagnostic. The \standardp/\mup control
disables depth scaling at the reference width. Workers average parameters and both \adamw moments at synchronization,
retaining optimizer ages.

\begin{widefigure}[!htbp]
 \centering
 \includegraphics[width=0.80\textwidth]{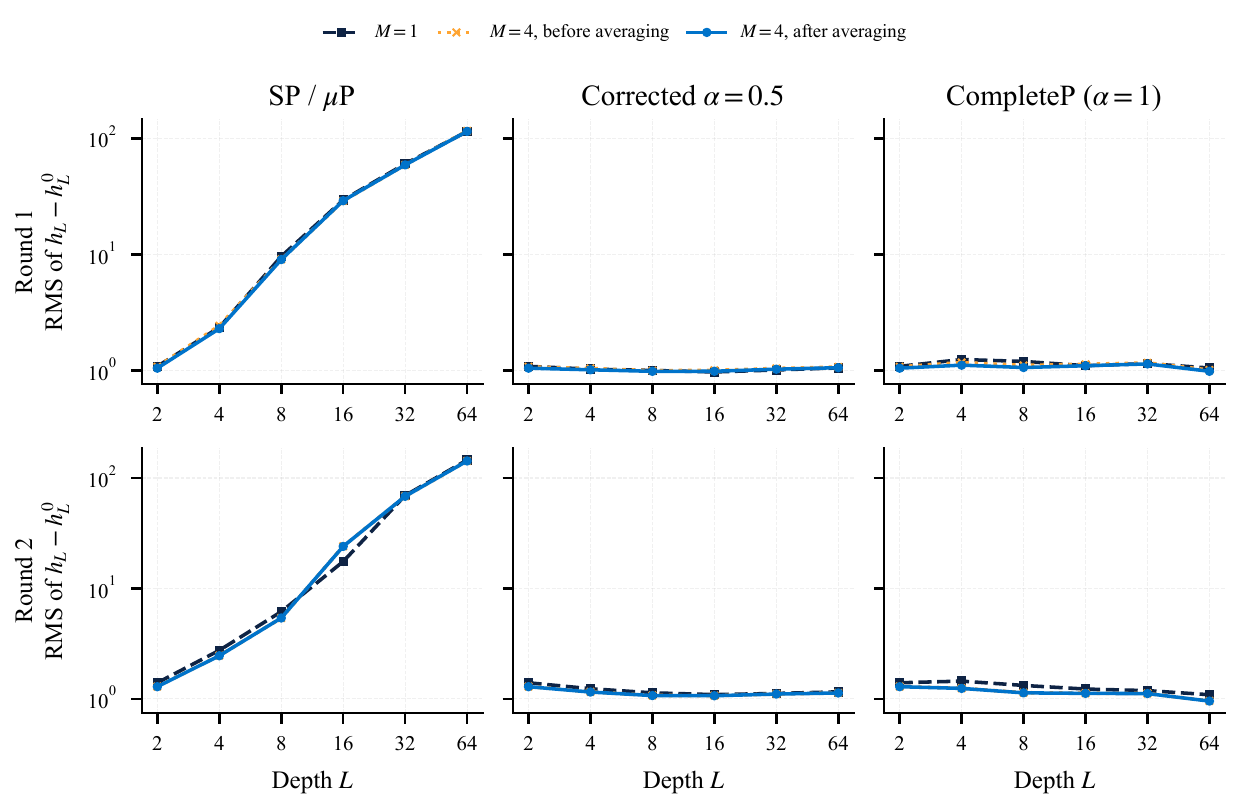}
 \caption{\textbf{Depth scaling controls feature movement through averaging.}
 Shakespeare transformer: RMS change in the last residual-block feature from initialization after \(32\) (top) and
 \(64\) (bottom) local steps. Curves compare one worker with four workers
 before and after parameter-and-moment averaging. Points are three-seed means.}
 \label{fig:shakespeare-feature-change}
\end{widefigure}

\paragraph{Measurements.}
To assess feature movement, let \(h_L\) denote the last residual-block
output before final LayerNorm.
We measure \(\operatorname{RMS}(h_L-h_L^0)\) on the probe sequences,
where \(h_L^0\) is the initial model's feature.
Before synchronization, we average the workers' measurements within each
seed, then across seeds. After synchronization, we measure the averaged model.
\Cref{fig:shakespeare-feature-change} compares this movement across depths.

\section{Mathematical Derivations}
\label{app:theory}
\setlength{\parskip}{3pt plus 1pt minus 1pt}
\setlength{\abovedisplayskip}{6pt plus 2pt minus 1pt}
\setlength{\belowdisplayskip}{6pt plus 2pt minus 1pt}
We provide the derivations supporting \cref{sec:new-distributed-training-scaling},
using the same notation and writing \(T=T_{\mathrm{inner}}\) for the total
number of inner steps.

\subsection{Width and depth under averaging}
\label{app:server-averaging-scaling}
We make the conditions for bounded, non-vanishing feature changes explicit,
building on \lordo's averaging argument~\citep{jovanovic2026lordo}.
As in \cref{sec:averaging-theory}, we evaluate changes at the shared
parameters \(\theta=\theta_t\).
Let \(\delta\theta^m=-\Delta_t^m\) be worker
\(m\)'s displacement, \(\bar\delta\theta\) its average across workers, and
\(u_m=J_{\ell,t}\delta\theta^m\) the first-order feature change. We use the
same feature normalization throughout.

For fixed \(M\) and \(K\), assume that there are constants
\(a,C,c>0\), independent of width and depth, such that
\[
 \begin{gathered}
 \max_m\|u_m\|\le C,\qquad \frac1M\sum_m\|u_m\|\ge a,\\
 \left\|\frac1M\sum_m u_m\right\|
 \ge\frac cM\sum_m\|u_m\|.
 \end{gathered}
\]
The first two conditions keep worker changes bounded with non-vanishing
mean magnitude. The third ensures that averaging retains a fixed fraction
of this magnitude, as measured in \cref{app:toy-averaging}.
Since \(J_{\ell,t}\bar\delta\theta=\frac{1}{M}\sum_m u_m\), the triangle
inequality gives an upper bound, while the last two conditions give a lower bound:
\begin{equation}
 ca\le\|J_{\ell,t}\bar\delta\theta\|\le C.
 \label{eq:app-feature-scale-bound}
\end{equation}
The actual feature change also includes the nonlinear remainder:
\[
 h_\ell(\theta+\bar\delta\theta)-h_\ell(\theta)
 =J_{\ell,t}\bar\delta\theta+r_\ell.
\]
If \(\|r_\ell\|\le\varepsilon\|J_{\ell,t}\bar\delta\theta\|\)
for a scale-independent \(0\le\varepsilon<1\), the triangle and reverse
triangle inequalities bound the actual change between
\((1-\varepsilon)ca\) and \((1+\varepsilon)C\).
Both bounds are independent of width and depth, giving the feature-scaling
result in \cref{eq:server-average-feature-scale}.

\subsection{Width and depth with outer optimizers}
\label{app:outer-scaling-contract}
Following \cref{sec:outer-model-scaling}, we ask whether scaling local
updates also scales the outer update. Suppose every pseudo-gradient is
multiplied by \(s>0\). For \sgd, \sgdm and \nesterov, keep the coefficients
fixed and either initialize momentum at zero or scale it by \(s\).
The outer update is linear in the full pseudo-gradient history, so it
also scales by \(s\), with no further correction.

For outer \adam, let \(m_t^{\mathrm{out}}\) and \(v_t^{\mathrm{out}}\) be
the moments of \(\Delta_t\). Scaling their initial values by \(s\) and
\(s^2\), while retaining the same moment coefficients and bias-correction
step counts, gives
\(m_t^{\mathrm{out}\prime}=sm_t^{\mathrm{out}}\) and
\(v_t^{\mathrm{out}\prime}=s^2v_t^{\mathrm{out}}\) at every step.
If epsilon also scales by \(s\), the normalized update satisfies
\[
 \frac{\widehat m_t^{\mathrm{out}\prime}}{\sqrt{\widehat v_t^{\mathrm{out}\prime}}+s\epsilon_{\mathrm{out}}}
 =\frac{\widehat m_t^{\mathrm{out}}}{\sqrt{\widehat v_t^{\mathrm{out}}}+\epsilon_{\mathrm{out}}},
\]
where \(\widehat m,\widehat v\) denote the bias-corrected moments, and denominators are assumed positive.
Normalization therefore removes the scale factor from each coordinate.
To recover the required update factor \(s_{\eta,\ell}\) for parameter
group \(\ell\), we scale the outer learning rate by that factor and
epsilon by the pseudo-gradient factor \(s_{\Delta,\ell}\).
For reference coefficients
\(\kappa_\gamma,\kappa_\epsilon\), this gives
\(\gamma_\ell=\kappa_\gamma s_{\eta,\ell}\) and
\(\epsilon_{\mathrm{out},\ell}=\kappa_\epsilon s_{\Delta,\ell}\).
Our width/depth prescription uses \(s_{\Delta,\ell}=s_{\eta,\ell}\).
This gives the outer-\adam correction in \cref{sec:outer-model-scaling},
assuming the pseudo-gradient histories scale uniformly across model sizes.

\subsection{Global and worker-local batch sizes}
\label{app:distributed-unit-scaling}
\paragraph{Larger local batches.}
\label{app:distributed-sde}
\label{app:rq2-sde-audit}
The argument in \cref{sec:worker-batch-duration-theory} uses the \adam SDE
to motivate the learning-rate rule \(\eta'=\sqrt\kappa\,\eta\) when
\(b'=\kappa b\). This choice preserves the local noise parameter
\(\eta\sigma_b\) and effective training horizon \(T\eta^2\) at fixed tokens.
This uses \(\sigma_b\propto b^{-1/2}\) at a common parameter state.
The full SDE prescription also rescales \(\beta_1,\beta_2\) and epsilon
\citep{malladi2022sde,mlodozeniec2025completedp}.
RQ2 asks whether this learning-rate rule remains effective while retaining
the reference optimizer configuration. We therefore keep the moment
coefficients, epsilon and synchronization interval fixed during batch
sweeps (\cref{sec:experimental-setup}). This holds the moment-averaging
windows and synchronization cadence fixed in local steps. At each batch
size, the learning-rate sweep then isolates the effect of the rate
prescription from changes to those settings.

\paragraph{Additional workers.}
\label{app:rq2-second-moment}
Increasing the worker count and increasing the local batch also affect
\adam's second moment differently. To see this, compare averaging local
second moments with updating a second moment from the averaged gradient
\citep{belilovsky2026replica}.
Consider one coordinate and one update from shared parameters and a common
second moment \(v\). For local gradients \(g^m\), write
\[
 \begin{gathered}
 \bar g=\frac{1}{M}\sum_m g^m,
 \qquad \bar v^+=\frac{1}{M}\sum_m v^{m,+},\\
 v^{m,+}=\beta_2v+(1-\beta_2)(g^m)^2,
 \qquad v_{\mathrm{DDP}}^+=\beta_2v+(1-\beta_2)\bar g^2.
 \end{gathered}
\]
Subtracting the two second-moment updates gives
\[
 \bar v^+-v_{\mathrm{DDP}}^+
 =\frac{1-\beta_2}{M}\sum_m(g^m-\bar g)^2.
\]
For IID gradients with variance \(\frac{\sigma^2}{b}\), this gives
\begin{equation}
 \mathbb E[\bar v^+-v_{\mathrm{DDP}}^+]
 =(1-\beta_2)\frac{\sigma^2}{b}\left(1-\frac1M\right).
 \label{eq:rq2-second-moment-gap}
\end{equation}
Thus, larger local batches reduce this difference, whereas adding workers
at fixed local batch increases it towards a finite limit. Averaging the
worker moments after local updates does not recover the second moment
computed by \ddp at global batch \(Mb\). This explains why the global
batch alone does not determine local \adam's normalization.

\Needspace*{10\baselineskip}
\subsection{Transfer across local steps}
\label{app:local-step-scaling}
\label{app:cadence-round-argument}
Changing \(K\) preserves the total step-size sum in \cref{sec:cadence-theory},
but changes movement between synchronizations. For local updates
\(-\eta u_k^m\), with \(\eta\ge0\) and zero weight decay, one round
of outer \sgd at rate \(\gamma\) gives
\begin{equation}
 \theta_{\mathrm{next}}-\theta_0
 =-\frac{\gamma\eta}{M}\sum_{m=1}^M\sum_{k=0}^{K-1}u_k^m.
 \label{eq:inner-round-displacement}
\end{equation}
If \(\|u_k^m\|\le G\) uniformly over the compared runs, then
\(\|\theta_k^m-\theta_0\|\le k\eta G\) and
\(\|\theta_{\mathrm{next}}-\theta_0\|\le|\gamma|K\eta G\).
At fixed learning rates, both bounds grow linearly with \(K\) and quantify
possible worker drift.

\section{Extended Related Work}
\label{app:related-work}

\paragraph{Communication and execution.}
Infrequent synchronization can still incur high communication volume and blocking.
\streamingdiloco~\citep{douillard2025streaming} overlaps communication with
training, while \sparseloco~\citep{sarfi2025sparseloco} compresses updates.
\asyncdiloco~\citep{liu2024asynchronous} addresses worker heterogeneity,
and \decoupleddiloco~\citep{douillard2026decoupled} tolerates delays and failures.
\asyncmesh~\citep{ajanthan2026asyncmesh} corrects for staleness, while
\factoredgossipdiloco~\citep{koneputugodage2026factored} reduces blocking through
approximate synchronization. These methods are orthogonal to our paper as they address the cost and scheduling
of synchronization, where we consider how hyperparameters should transfer
when model size and training configuration change.

\paragraph{Optimizer state and memory.}
Local training must also balance shared optimizer history against communication
and memory costs. \diloco~\citep{douillard2023diloco} retains local states,
whereas \photon~\citep{sani2025photon} resets them with intermittent clients. However, resetting \adam's second moment harms translation
performance in \slowmo~\citep{wang2020slowmo}.
\desloc~\citep{iacob2025desloc} reduces communication by synchronizing
parameters and moments independently. \mtdao~\citep{iacob2025mtdao} uses
fast and slow momenta to retain shared information during local training.
\lordo~\citep{jovanovic2026lordo} reduces optimizer-state memory through low-rank
updates, while \fomoe~\citep{sani2026fomoe} distributes experts across workers.
These choices determine how optimizer history is retained and distributed.
We study hyperparameter transfer under a specified optimizer and
synchronization policy.

\paragraph{Batch and optimizer scaling.}
Batch-size gains eventually saturate~\citep{mccandlish2018batch,zhang2025critical}.
The \adam SDE~\citep{malladi2022sde} and
\completedp~\citep{mlodozeniec2025completedp} motivate adjusting training
settings with batch size. We examine how this extends to local workers.
Scaling also depends on the optimizer. Matrix-preconditioned methods such as
\muon~\citep{liu2025muon} require appropriate model-size
scaling~\citep{qiu2025matrixpreconditioned}. \muloco~\citep{therien2025muloco}
applies \muon locally, and \citet{belilovsky2026replica} link inner
preconditioning to replica scaling. Our question is how the inner rule and
outer update jointly determine transfer in local-update training. We assess
this across training budgets and model scales~\citep{semenov2025benchmarking,wen2026fantastic}.

\section{Limitations}
\label{app:limitations}
In our experiments, we focus on the use of inner \adam given that the zero-shot hyperparameter transfer frameworks were developed for this setting. However, we have not evaluated transfer with
other inner--outer optimizer combinations. Further to this, we also assume synchronous
updates, leaving open how asynchrony and update staleness interact with
the transfer rules and optimizer-state synchronization. Finally, although our
formulation includes token-budget scaling, our batch sweeps hold tokens
fixed, so we leave its empirical evaluation to future work.

\section{AI Usage}
\label{app:ai-usage}
Throughout this work, we have used AI tools to assist with writing and preparing the manuscript. This has
included revising prose, reducing repetition, formatting LaTeX, and
developing plotting code. We also used LLMs as coding assistants. Additionally, we used LLMs to find relevant literature,
check claims against supporting sources, and help verify the correctness and clarity of our
mathematical arguments. All reported experimental results come from the training runs
described in this paper. We remain responsible for the paper and its conclusions.

\end{document}